\documentclass{article} 
\usepackage{iclr2027_conference,times}

\usepackage{amsmath,amsfonts,bm}

\def\eqref#1{equation~\ref{#1}}

\def\1{\bm{1}}

\DeclareMathAlphabet{\mathsfit}{\encodingdefault}{\sfdefault}{m}{sl}
\SetMathAlphabet{\mathsfit}{bold}{\encodingdefault}{\sfdefault}{bx}{n}

\usepackage{hyperref}
\usepackage{url}
\usepackage[utf8]{inputenc} 
\usepackage[T1]{fontenc}    
\usepackage{hyperref}       
\usepackage{url}            
\usepackage{booktabs}       
\usepackage{amsfonts}       
\usepackage{nicefrac}       
\usepackage{microtype}      
\usepackage{xcolor}         
\usepackage{amsmath}
\usepackage{amsfonts}
\usepackage{amsthm}
\usepackage{dsfont}
\usepackage{array}   
\usepackage[ruled,vlined,linesnumbered]{algorithm2e}
\usepackage{enumitem}
\usepackage{amssymb}

\usepackage{adjustbox}
\usepackage{booktabs}
\usepackage{threeparttable}
\usepackage[dvipsnames,table]{xcolor}
\usepackage{colortbl}
\usepackage{multirow}

\usepackage{tabularx}
\usepackage{wrapfig}
\usepackage{natbib}
\usepackage{graphicx}
\usepackage{subcaption} 
\usepackage{wrapfig}
\usepackage{hyperref}
\usepackage{todonotes} 
\usepackage[most]{tcolorbox}
\makeatother
\usepackage[most]{tcolorbox} 
\definecolor{myblue}{HTML}{499BC0}
\definecolor{myred}{HTML}{F78779}
\definecolor{myblueedge}{RGB}{40,60,130} 
\definecolor{myblueback}{RGB}{235,242,255} 
\definecolor{myblue1}{HTML}{015696}
\definecolor{myblueedge1}{HTML}{0B75B3}
\definecolor{ysdarkpurple}{HTML}{4E2399}
\definecolor{ysshallowpurple}{HTML}{E6DBFF}
\definecolor{ysdarkred}{HTML}{8c2824}
\definecolor{ysshallowred}{HTML}{F8D7D7}
\definecolor{ysdarkblue}{HTML}{005E99}
\definecolor{ysshallowblue}{HTML}{CCEBFF}
\definecolor{ysdarkgrey}{HTML}{333333}
\definecolor{ysshallowgrey}{HTML}{E5E5E5}
\definecolor{text_red}{HTML}{982B2D}
\definecolor{image_blue}{HTML}{012A61}
\definecolor{positive}{HTML}{008000} 
\definecolor{negative}{HTML}{FF0000}
\definecolor{blue4}{HTML}{335372}
\definecolor{red4}{HTML}{E25659}
\definecolor{morandiblue}{HTML}{E6ECF1}
\definecolor{morandiBack}{HTML}{EEF2F6}
\definecolor{morandiFrame}{HTML}{6F8AA3}
\definecolor{morandiTitle}{HTML}{3E5A75}
\definecolor{morandiVar}{HTML}{2F4A66}
\definecolor{morandiAccent}{HTML}{8FA8BD}

\definecolor{linkblue}{HTML}{2459A6}
\hypersetup{
    colorlinks=true,
    urlcolor=linkblue,
    citecolor=linkblue,
    linkcolor=linkblue
}

\newcommand{\promptvar}[1]{\textcolor{morandiVar}{<#1>}}

\newtcolorbox{findingbox}{
  enhanced,
  colback=myblueback,
  colframe=myblueedge,
  fontupper=\bfseries\itshape\normalsize,
  arc=1mm,
  boxrule=0.8pt,
  top=3mm,
  bottom=3mm,
  left=3mm,
  right=3mm,
  drop shadow=black!30!white,
}
\iclrfinalcopy

\usepackage{etoolbox}
\makeatletter
\patchcmd{\@maketitle}
  {\lhead{Published as a conference paper at ICLR 2027}}
  {\lhead{Preprint}}
  {}{}
\makeatother

\newcommand{\mybench}{TraceAV-Bench}
\title{TraceAV-Bench: Benchmarking Multi-Hop Trajectory Reasoning over Long Audio-Visual Videos}

\vspace{-2mm}
\author{%
Hengyi Feng\textsuperscript{1,2}\thanks{Equal contribution}\hspace{0.4em},%
\quad
Hao Liang\textsuperscript{2,4}\footnotemark[1]\hspace{0.5em}\thanks{Project Leader}\hspace{0.3em},%
\quad
Mingrui Chen\textsuperscript{3},%
\quad
Bohan Zeng\textsuperscript{2},%
\quad
Meiyi Qiang\textsuperscript{2},\\
\textbf{Zhengyang Zhao\textsuperscript{2},%
\quad
Zimo Meng\textsuperscript{2},%
\quad
Zeang Sheng\textsuperscript{2},%
\quad
Wentao Zhang\textsuperscript{2,4}}\thanks{Corresponding author}\\[0.8em]
\textsuperscript{1}University of Electronic Science and Technology of China\\
\textsuperscript{2}Peking University\\
\textsuperscript{3}Institute of Automation, Chinese Academy of Sciences\\
\textsuperscript{4}Zhongguancun Academy\\[0.3em]
\textbf{Project page:} \textbf{\small\url{https://heinz217.github.io/TraceAV-Bench-Page}}%
}
\vspace{-2mm}

\begin{document}

\maketitle

\begin{abstract}
Real-world audio-visual understanding requires chaining evidence that is sparse, temporally dispersed, and split across the visual and auditory streams, whereas existing benchmarks largely fail to evaluate this capability.
They restrict videos to short clips, isolate modalities, or reduce questions to one-hop perception.
We introduce \textbf{\mybench}, the first benchmark to jointly evaluate \textbf{\textit{multi-hop reasoning over long audio-visual trajectories}} and \textbf{\textit{multimodal hallucination robustness}}.
\mybench\ comprises 2{,}200 rigorously validated multiple-choice questions over 578 long videos, totaling 339.5 hours, spanning 4 evaluation dimensions and 15 sub-tasks.
Each question is grounded in an explicit multi-hop evidence trajectory that averages 3.68 hops across a 15.1-minute temporal span.
The dataset is built by a three-step pipeline followed by a strict quality assurance process.
Evaluation of multiple representative OmniLLMs on \mybench\ reveals that the benchmark poses a persistent challenge across all models, with the strongest closed-source model (Gemini~3.1~Pro) reaching only 68.29\% on general tasks, and the best open-source model (Ming-Flash-Omni-2.0) reaching 51.70\%, leaving substantial headroom.
Further analysis shows that current models struggle to gather and combine evidence across long audio-visual videos, with performance varying across modalities and reasoning tasks.
We hope \mybench\ will facilitate progress toward reliable long-form audio-visual reasoning in OmniLLMs.
\end{abstract}

\section{Introduction}

The rapid advancement of Multimodal Large Language Models (MLLMs)~\cite{bai2025qwen3, yao2024minicpmv, yin2024survey, team2025kimivl} has broadened the perceptual horizon of language models, enabling them to process visual~\cite{zhu2025internvl3, steiner2024paligemma2} and auditory~\cite{team2026qwen35asr, shi2026qwen3asrtechnicalreport} information far beyond the boundaries of language alone.
Building upon this foundation, Omnimodal Large Language Models (OmniLLMs)~\cite{team2026qwen35omni, ai2025ming, zhao2025humanomni, team2025longcat, tang2025video, li2024baichuan, comanici2025gemini} have recently emerged, designed to jointly perceive and reason over 
real-world text, vision, and audio within a unified framework, achieving competitive performance across tasks such as
audio-visual understanding and cross-modal reasoning~\cite{han2024onellm, wang2025omnimmi, li2026omnigaia}.
This capability is fundamental for interacting with the physical world in a human-like manner, in which information from different modalities is simultaneously integrated.

However, despite these promising results, bringing OmniLLMs into genuine real-world utility reveals several unresolved challenges.
\textit{\textbf{1) Long-form multi-hop trajectory reasoning}} poses the most critical bottleneck~\cite{tao2026lvomnibench, wang2025lvbench, fu2025videomme, chen2026futureomni}.
While OmniLLMs can handle simple, short-clip perception tasks, real-world scenarios often demand chaining clues scattered across tens of minutes of continuous audio-visual content.
This requires reasoning from a \textit{trajectory} of temporally dispersed evidence spanning multiple events, rather than a single moment.
\textit{\textbf{2) Cross-modal information fusion}} remains equally unresolved~\cite{zhou2025daily, li2025omnivideobench}. 
A model that can process audio and visual streams does not necessarily know how to jointly leverage their complementary information under complex conditions, particularly when evidence in one modality is only interpretable in the context of the other.
\textit{\textbf{3) Multimodal hallucination}} further compounds these difficulties~\cite{bai2024hallucination, sung2024avhbench, xing2026learning}.
When jointly processing mixed audio-visual inputs, OmniLLMs sometimes generate responses that are inconsistent with the input or not grounded in the observed content, raising serious reliability concerns.
Critically, these challenges remain not only largely unsolved, but also insufficiently studied.
Existing benchmarks mostly operate on short video clips of at most a few minutes, evaluate visual and audio modalities in isolation, or restrict questions to shallow single-hop inference, failing to comprehensively evaluate these challenges~\cite{li2024omnibench, gong2024av}.

\begin{figure}[t]
    \centering
    \includegraphics[width=\textwidth]{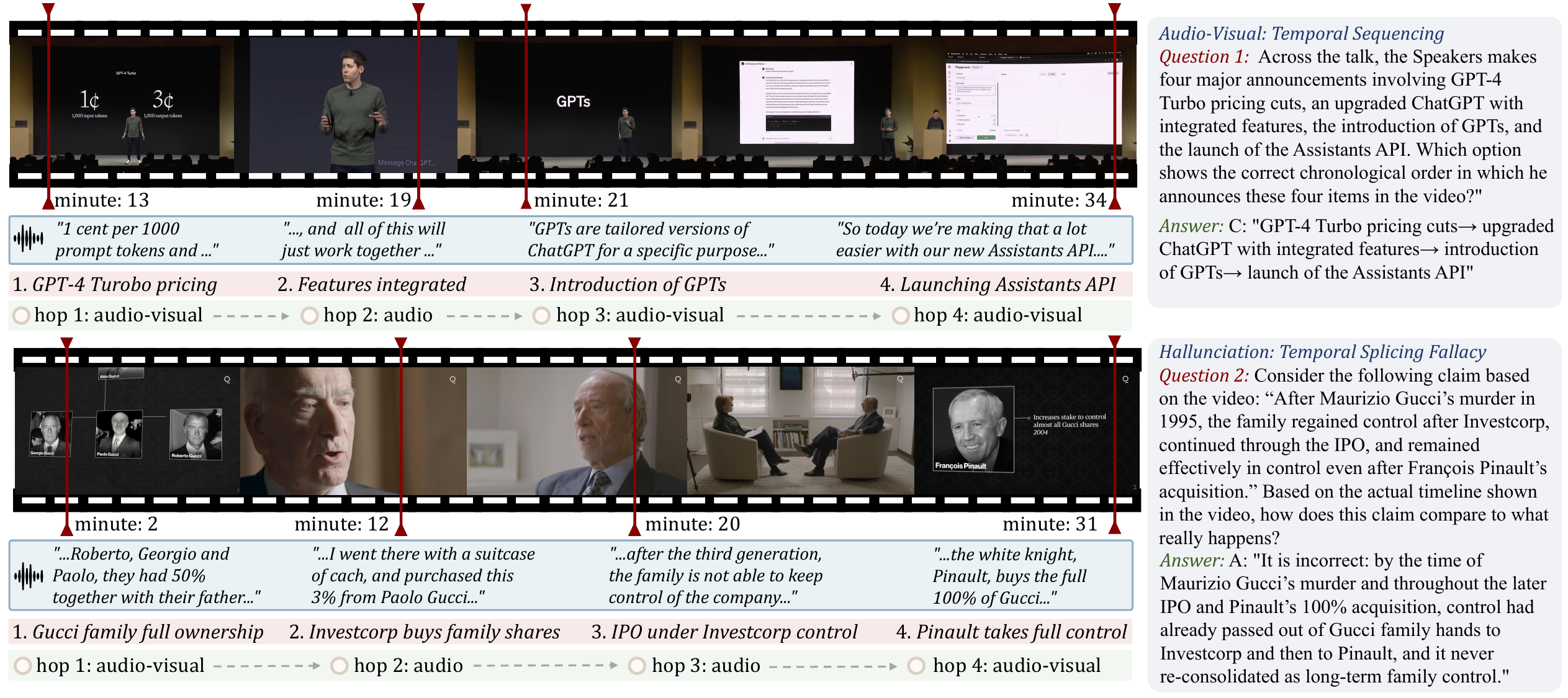}
    \vspace{-5mm}
    \caption{Illustrative examples from \mybench, each grounded in an explicit multi-hop evidence trajectory whose hops are tagged with their source modality. The first example (\textit{Temporal Sequencing}, AVR) requires chronologically ordering four events by chaining speech and on-screen cues; the second (\textit{Temporal Splicing Fallacy}, MH) requires rejecting a fabricated narrative that splices temporally isolated events into a false ownership timeline. Both showcase the main idea of \mybench: composing evidence from temporally dispersed, cross-modal clues over long videos.}
    \label{fig:examples}
    \vspace{-3mm}
\end{figure}

To this end, we introduce \textbf{\mybench}, a comprehensive benchmark designed to evaluate OmniLLMs on multi-hop reasoning over long audio-visual trajectories.
\mybench\ is built upon 578 long videos totaling 339.5 hours, spanning diverse genres and multiple languages, and it is, to the best of our knowledge, the \textit{\textbf{first}} benchmark to simultaneously require multi-hop trajectory reasoning and assess multimodal hallucination robustness in the context of long audio-visual content.
Via a three-step data synthesis pipeline, we construct 2,200 rigorously validated multiple-choice questions (MCQs) across 4 evaluation dimensions and 15 sub-tasks, covering \textbf{\textit{Audio-Visual Joint Reasoning}}, \textbf{\textit{Visual-Centric Reasoning}}, \textbf{\textit{Audio-Centric Reasoning}}, and \textbf{\textit{Multimodal Hallucination}}.
Every question in \mybench\ is grounded in an explicitly annotated multi-hop evidence trajectory (as showcased in Figure~\ref{fig:examples}), where each reasoning chain spans at least two temporally non-adjacent events, and averages 3.68 hops across a temporal span of 15.1 minutes.

Our main contributions can be summarized as follows:
\begin{itemize}[leftmargin=*, itemsep=2pt]
\item \textit{\textbf{The \mybench\ Benchmark.}}
We present a large-scale, carefully curated benchmark that extends the evaluation frontier for OmniLLMs over long multi-hop trajectories, providing the research community with a tool to rigorously diagnose model capabilities beyond existing benchmarks.

\item \textit{\textbf{A Data Construction Pipeline.}}  
We propose a three-step pipeline that decouples visual captioning, asynchronous audio-visual fusion, and LLM-based question generation to organize asynchronous evidence from long videos into logically coherent, trajectory-grounded MCQs at scale.  
A rigorous multi-stage quality assurance process further ensures that every retained question is of high quality.

\item \textit{\textbf{Comprehensive Empirical Analysis with Key Findings.}}
We benchmark state-of-the-art open-source and closed-source OmniLLMs and MLLMs, and find that all evaluated models struggle on \mybench.
When answering multi-hop questions over long videos, these models have difficulty gathering and combining the required audio and visual evidence from different parts of the video to reach the correct answer.
Our analysis further shows that these limitations vary across modalities and reasoning operations.
\end{itemize}

\begin{table}[t]
\vspace{-1mm}
\caption{Comparison of \mybench\ with existing audio-visual benchmarks. V, I, A: video, image, audio. Anno.: M = manual, A\&M = automatic + manual.}
\vspace{-2mm}
\label{tab:comparison}
\centering
\small
\setlength{\tabcolsep}{4pt}
\renewcommand{\arraystretch}{1.05}
\resizebox{0.98\linewidth}{!}{
\begin{tabular}{lcccccccc}
\toprule
\textbf{Benchmark} & \textbf{Modality} & \textbf{Duration} & \textbf{Avg.\ Length} & \textbf{Multi-hop} & \textbf{Multi-domain} & \textbf{Hallucination} & \textbf{Anno.} & \textbf{\#Q} \\
\midrule
\rowcolor{morandiblue}\multicolumn{9}{l}{\textit{\textbf{Short-clip benchmarks ($\leq 60$ s)}}} \\
\midrule
AVQA~\cite{yang2022avqa}            & V+A  & $\sim$10 s       & $\sim$10 s   & \texttimes & \texttimes & \texttimes & M    & 57,335 \\
Music-AVQA~\cite{li2022learning}      & V+A  & $\sim$60 s       & $\sim$60 s   & \texttimes & \texttimes & \texttimes & M    & 45,867 \\
AVTRUSTBENCH~\cite{chowdhury2025avtrustbench}    & V+A  & 10--60 s         & -            & \texttimes & \checkmark & \checkmark & A\&M & 600 K  \\
AV-Odyssey~\cite{gong2024av}      & I+A  & /                & /            & \texttimes & \checkmark & \texttimes & M    & 4,555  \\
AVHBench~\cite{sung2024avhbench}        & V+A  & $\sim$10 s       & $\sim$10 s   & \texttimes & \checkmark & \checkmark & A\&M & 5,302  \\
OmniBench~\cite{li2024omnibench}       & I+A  & -                & -            & \texttimes & \checkmark & \texttimes & M    & 1,142  \\
Daily-Omni~\cite{zhou2025daily}      & V+A  & 30--60 s         & 42.8 s       & \texttimes & \texttimes & \texttimes & A\&M & 1,197  \\
\midrule
\rowcolor{morandiblue}\multicolumn{9}{l}{\textit{\textbf{Long-form video benchmarks ($>60$ s)}}} \\
\midrule
WorldSense~\cite{hong2025worldsense}      & V+A  & 15--656 s        & 141.1 s      & \texttimes & \checkmark & \texttimes & M    & 3,172  \\
JointAVBench~\cite{chao2025jointavbench}    & V+A  & -                & 97.2 s       & \texttimes & \checkmark & \texttimes & A\&M & 2,853  \\
OmniVideoBench~\cite{li2025omnivideobench}  & V+A  & 4--1955 s        & 384.2 s      & \texttimes & \checkmark & \texttimes & M    & 1,500  \\
LVOmniBench~\cite{tao2026lvomnibench}     & V+A  & 613--5482 s      & 2{,}070 s    & \texttimes & \checkmark & \texttimes & M    & 1,014  \\
\midrule
\textbf{\mybench\ (Ours)} & \textbf{V+A} & \textbf{606--8394 s}\,\textcolor{positive}{\scriptsize\textbf{(Ultra-long)}} & \textbf{2{,}112 s} & \textcolor{positive}{\boldmath$\checkmark$} & \textcolor{positive}{\boldmath$\checkmark$} & \textcolor{positive}{\boldmath$\checkmark$} & \textbf{A\&M} & \textbf{2,200} \\
\bottomrule
\end{tabular}
}
\vspace{-2mm}
\end{table}

\section{Related Work}

\subsection{Omnimodal Large Language Models}

Real-world perception is inherently multi-sensory, and closing the gap between human and machine understanding requires models that can handle vision, audio, and language.
The research community has broadened the perceptual scope of LLMs, giving rise to Vision-Language Models (VLMs)~\cite{wang2025internvl35, bai2025qwen3, team2026qwen35omni, marafioti2025smolvlm, guo2025seed15vl, liu2025nvila, xiaomi2025mimo}, Audio-Language Models (ALMs)~\cite{tseng2025taste, ding2025kimiaudio, ghosh2025audioflamingo, chu2024qwen2audio, arora2025landscape}, and, more recently, OmniLLMs~\cite{liu2025ola, luo2025next, fu2024vita, guo2025m2, team2026qwen35omni, tian2026emoomni, ding2026omnisift} capable of processing text, images, videos, and audio within a unified framework.
Both open-source models~\cite{ye2025omnivinci, ai2025ming, zhao2025humanomni, tang2025video, xu2025qwen3, xu2025qwen25omnitechnicalreport} and closed-source models such as the Gemini series~\cite{comanici2025gemini} have demonstrated strong performance across audio-visual understanding tasks.
Meanwhile, multimodal research has continued to expand toward personalized streaming video understanding~\cite{zheng2026pearl}, unified personalized understanding and generation~\cite{an2025unictokens}, parameter-efficient multi-concept personalization~\cite{an2024mc}, recognition and understanding of images and videos~\cite{lin2025perceiveanythingrecognizeexplain}, and advances in visual understanding and generation through new architectures~\cite{yan2026pixel} and optimization strategies~\cite{yan2026do, yan2025entropy, yan2026litexplorer, yan2026multitune}.
Despite this rapid progress, rigorous evaluation of OmniLLMs on extended audio-visual content remains scarce.

\subsection{MLLM Benchmarks}

MLLM benchmarks have evolved with model capabilities, progressing from modality-isolated suites for image~\cite{liu2024mmbench, yue2024mmmu, yu2023mm} and audio~\cite{sakshi2024mmau, wang2025audiobench, wang2025mmsu, wang2025they} understanding to specialized axes such as fine-grained perception~\cite{ghosh2026understanding, qiang2025ver, weng2026oddgridbench} and multimodal reasoning~\cite{ying2024mmt, yao2025mmreason, du2025easy, li2026mmr, qian2025prism, fu2026videommev2, rawal2024cinepile}.
Recent evaluation suites also probe generative fluid intelligence in unified multimodal models, emphasizing contextual pattern induction, constraint execution, and adaptation beyond conventional knowledge-centric evaluation~\cite{an2026genius}.
The rise of OmniLLMs has spurred a rich set of audio-visual benchmarks~\cite{zhou2025daily, li2024omnibench, chen2026futureomni, cheng2025video, chen2025uno, gong2024av, zhang2025omnieval}. These benchmarks have made important progress in broadening evaluation from isolated perception to joint audio-visual understanding. 
They provide a strong foundation for studying the capabilities of OmniLLMs.
Several recent benchmarks have also extended evaluation to long-form videos~\cite{tao2026lvomnibench, li2025omnivideobench, han2025longinsightbench}. 
These studies demonstrate the value of realistic extended content and reveal the difficulty of maintaining reliable multimodal understanding over time. 
Building on these contributions, \mybench\ focuses on a complementary setting in which questions require multi-hop reasoning across temporally distant audio-visual evidence. 
It also evaluates multimodal hallucination robustness under the long-form setting, as summarized in Table~\ref{tab:comparison}.

\section{\mybench}
\label{sect:bench}

\subsection{Dataset Collection}
To evaluate the multi-hop reasoning capabilities of OmniLLMs over long trajectories in an audio-visual context, we define eligible long videos as those exceeding \textbf{10 minutes}, possessing semantically diverse and rich scene transitions, and containing aligned, meaningful information in both visual and audio modalities.

\begin{figure}[t]
    \centering
    \resizebox{0.91\textwidth}{!}{ 
    \begin{minipage}[t]{0.70\textwidth}
        \vspace{0pt} 
        \centering
        \includegraphics[width=\textwidth]{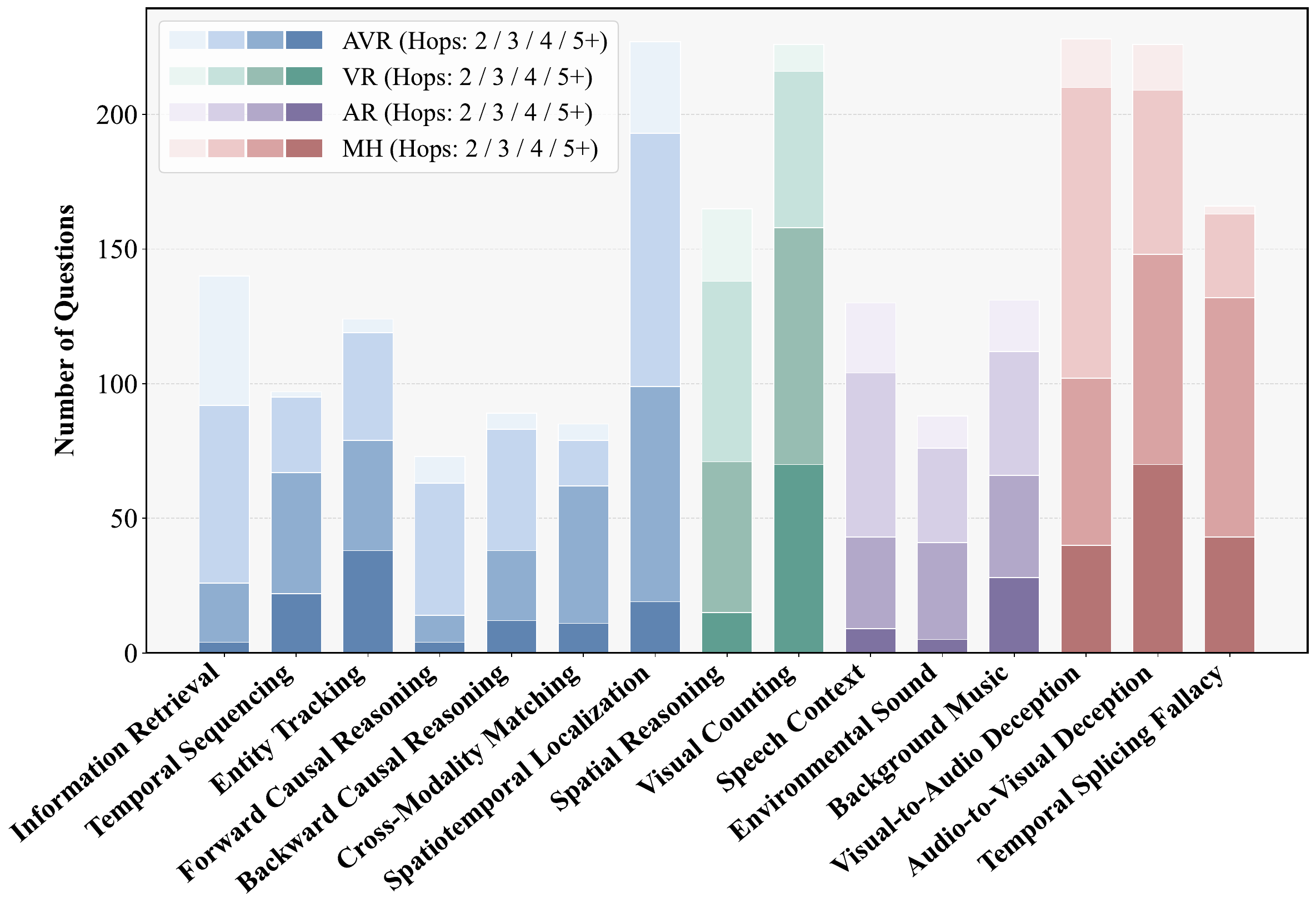}
    \end{minipage}
    \hfill 
    \begin{minipage}[t]{0.29\textwidth}
        \vspace{0pt} 
        \centering
        \includegraphics[width=0.8\textwidth]{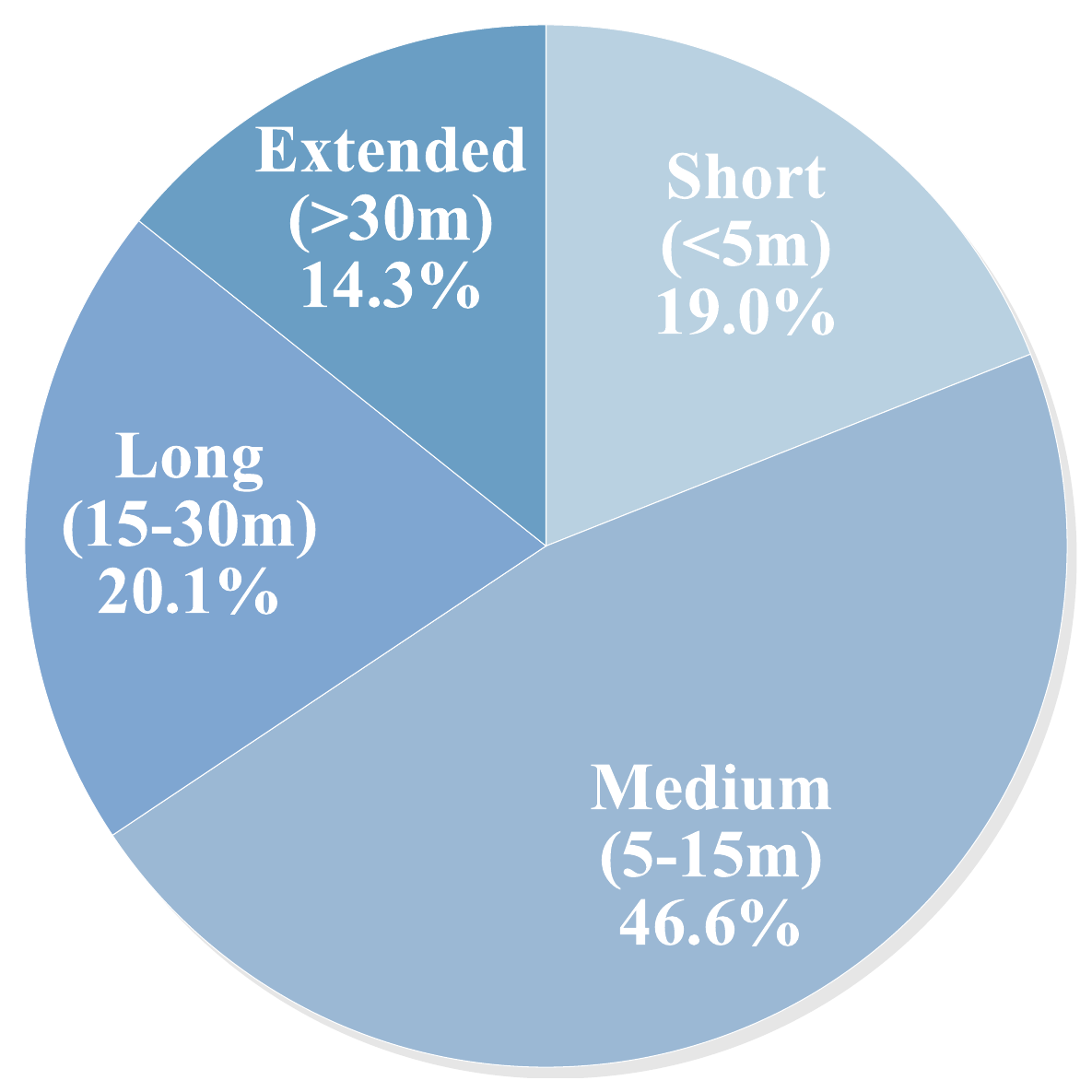}
        \includegraphics[width=0.8\textwidth]{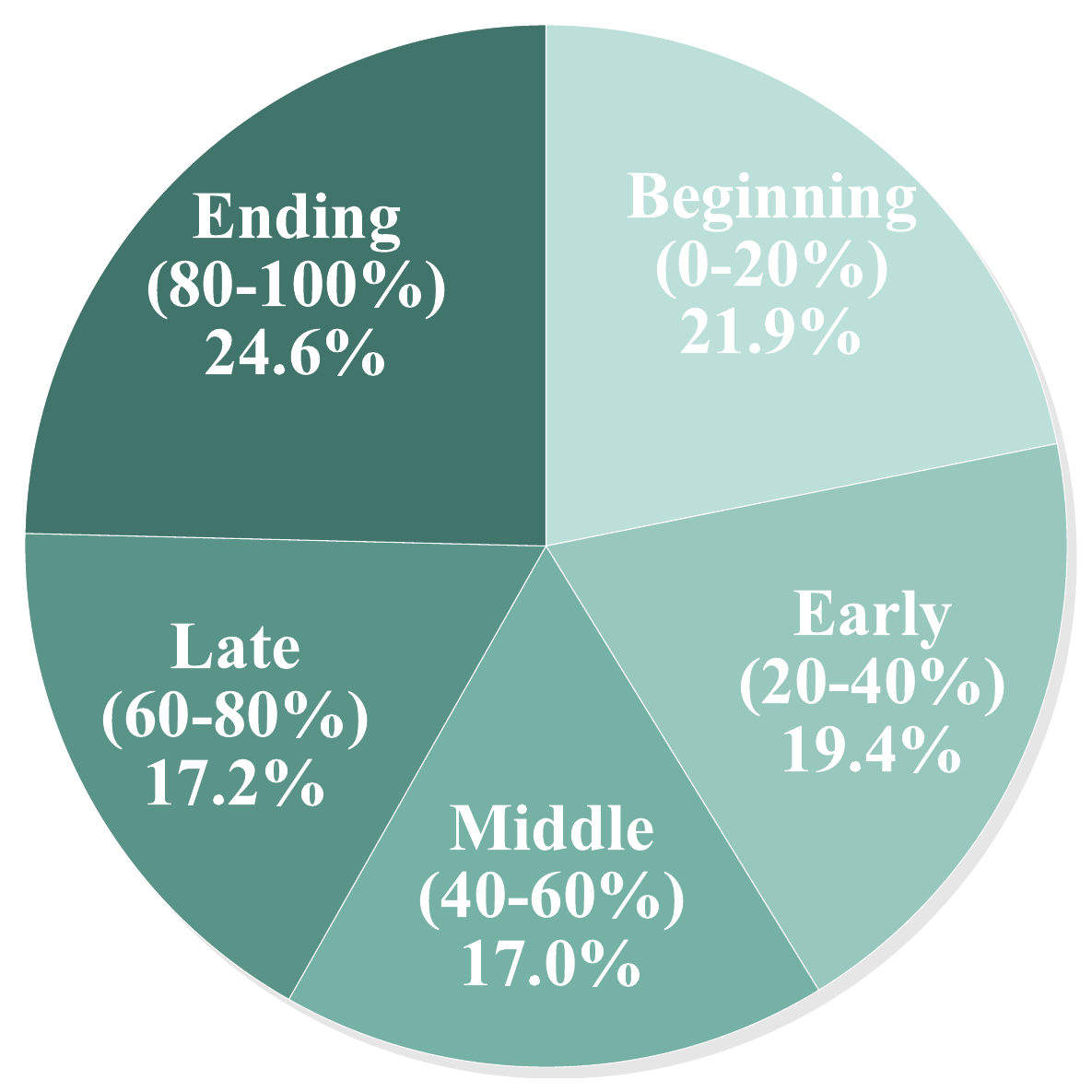}
    \end{minipage}
    }
    \vspace{-3mm}
    \label{fig:stats}
    \caption{Statistics analysis of \mybench. \textit{left}: per-sub-task question counts by evaluation dimension (AVR, VR, AR, MH), stacked by hop count ($2 / 3 / 4 / 5+$). \textit{right top}: distribution of video durations. \textit{right bottom}: distribution of the position for all evidence hops along the video timeline.}
    \vspace{-6mm}
\end{figure}

\begin{wrapfigure}{r}{0.40\textwidth} 
    \centering
    \vspace{-7.5mm}
    \includegraphics[width=0.36\textwidth]{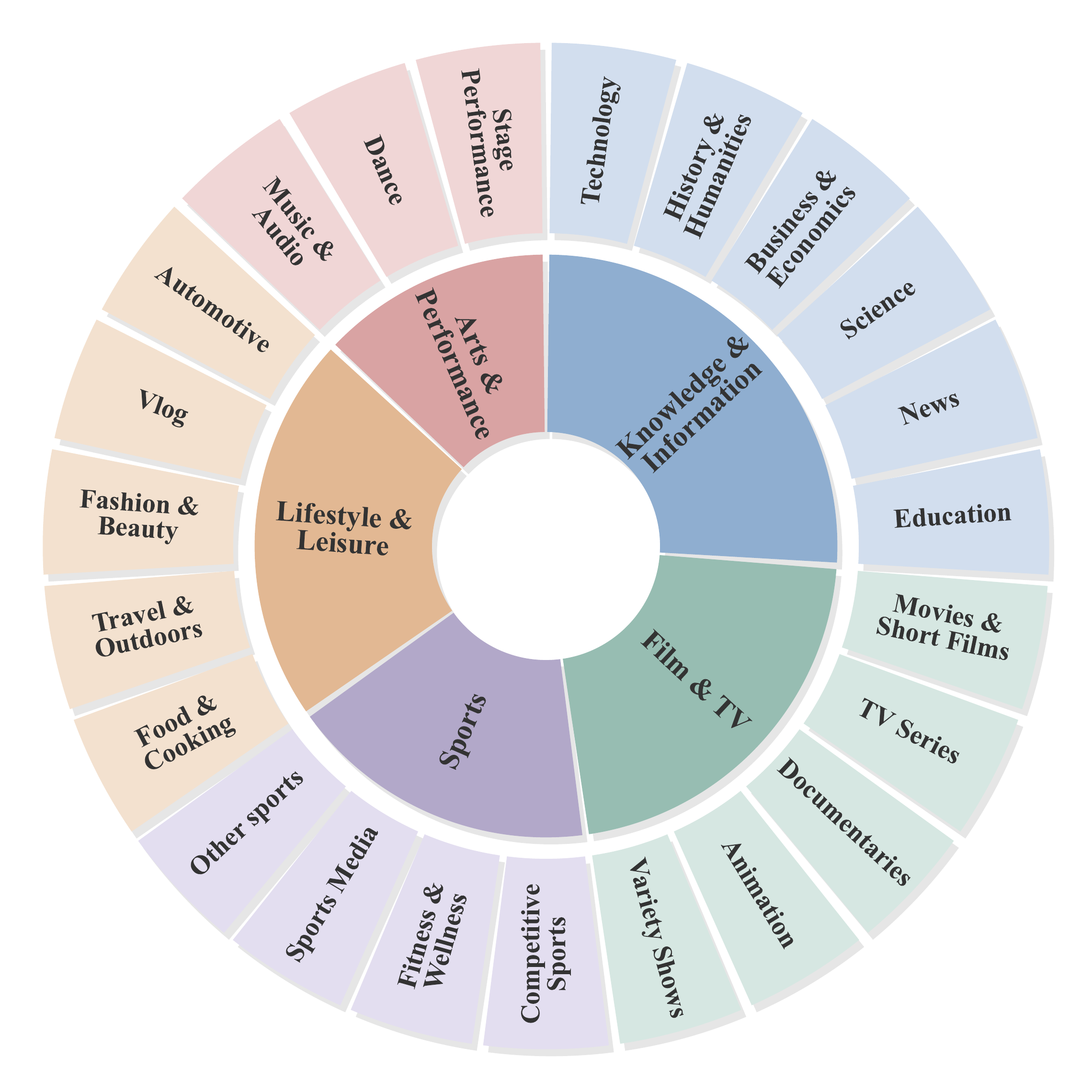}
    \vspace{-2.5mm} 
    \caption{Video category distribution.}
    \label{fig:video_categories}
    \vspace{-8mm} 
\end{wrapfigure}

\textbf{Video Collection and Deduplication.} 
Our initial video corpus is mainly sourced from three publicly available video understanding benchmarks: OmniVideoBench~\cite{li2025omnivideobench}, LVBench~\cite{wang2025lvbench} and VideoMME~\cite{fu2025videomme}, where YouTube is their primary source, there is a natural risk of data overlap. 
To ensure the uniqueness of our benchmark, we apply a rigorous deduplication process based on video metadata and content hashing, yielding a diverse initial pool of candidate videos.

\textbf{Video Quality Filtering.} We further process the candidate videos through a strict filtering pipeline with the following criteria with a half-automated manner:
\vspace{-2mm}
\begin{itemize}[leftmargin=*, itemsep=0pt]
    \item \textit{\textbf{Visual Dynamics and Event Density:}} 
    Videos must exhibit sufficient visual dynamics, with frequent scene transitions and a steady stream of distinguishable events. 
    Therefore, videos with fewer than 3 scene transitions are directly discarded.
    
    \item \textit{\textbf{Modality Completeness:}} 
    \mybench\ explicitly evaluates the joint understanding of audio and visual streams. 
    Thus, videos without an audio track or with a mean volume below -50 dB (indicating silence or negligible background noise) are discarded. 
    
    \item \textit{\textbf{Multi-hop Reasoning Potential:} }
    Our annotators manually review the videos to assess their potential for extracting multi-hop questions. 
    Specifically, videos are retained only if they exhibit high intrinsic complexity, such as multiple interacting entities and linked events. 
\end{itemize}
\vspace{-2mm}

After executing this filtering process, we finalized a high-quality collection of \textbf{578} videos, yielding a total of \textbf{339.5} hours of continuous audio-visual content ready for complex multi-hop construction. 
As shown in Figure~\ref{fig:video_categories}, the retained videos span a wide range of genres, ensuring broad topical diversity. 
Notably, the corpus is also multilingual, covering English, Chinese, and other major languages.

\subsection{Task Definition}

We meticulously design 15 distinct sub-tasks (with detailed definitions in Appendix~\ref{sec:appendix_task_def}).
These tasks are categorized into four core evaluation dimensions, briefly described as follows:
\vspace{-1mm}
\begin{itemize}[leftmargin=*, itemsep=2pt]
    \item \textit{\textbf{Audio-Visual Joint Reasoning (AVR):}} 
    This category forms the core of \mybench, evaluating the model's ability to chain clues across both visual and auditory streams over long temporal spans. 
    It includes seven sub-tasks: \textit{Information Retrieval (IR)}, \textit{Temporal Sequencing (TS)}, \textit{Entity Tracking (ET)}, \textit{Forward Causal Reasoning (FCR)}, \textit{Backward Causal Reasoning (BCR)}, \textit{Cross-Modality Matching (CMM)}, and \textit{Spatiotemporal Localization (SL)}.

    \item \textit{\textbf{Visual-Centric Reasoning (VR):}} 
    Rather than joint reasoning, this dimension isolates the visual modality to evaluate how well the model maintains its visual reasoning capabilities when faced with long video inputs. 
    It comprises \textit{Spatial Reasoning (SR)} and \textit{Visual Counting (VC)}.
    
    \item \textit{\textbf{Audio-Centric Reasoning (AR):} }
    Complementary to VR, this dimension aims to assess the model's auditory capabilities under long-video settings. 
    It encompasses examination across three distinct aspects: \textit{Speech Context (SC)}, \textit{Environmental Sound (ES)}, and \textit{Background Music (BM)}.
    
    \item \textit{\textbf{Multimodal Hallucination (MH):}} 
    This category specifically investigates whether the model exhibits hallucinations in audio-visual contexts. It tests model robustness through \textit{Visual-to-Audio Deception (V2A)}, \textit{Audio-to-Visual Deception (A2V)}, and \textit{Temporal Splicing Fallacy (TSF)}.
\end{itemize}
\vspace{-2mm} 

\subsection{Data Construction Pipeline}
\vspace{-1mm}
\label{sect:pipeline}

To construct high-quality multi-hop questions over long audio-visual videos at scale, we develop a decoupled three-step pipeline that converts each video into a dense textual event catalog for trajectory-grounded question generation.
LLMs have also been employed as dataset analysts to uncover latent subpopulation structure~\cite{luo2024llm}, demonstrating their broader utility in data-centric workflows.

\textbf{Step 1: Minute-Level Visual Captioning.}
To obtain fine-grained visual descriptions throughout long videos, we divide each video into 1-minute clips and sequentially process them with Qwen3-VL-32B-Instruct~\cite{bai2025qwen3}. 
At this stage, we isolate the visual stream to avoid cross-modal interference and hallucination during initial content extraction~\cite{xie2026maverix, sung2024avhbench}. 
To preserve identity consistency across temporally segmented clips, we maintain an \textit{Entity Cache} that records previously identified persons, objects, locations, and their evolving states.

\textbf{Step 2: Audio-Visual Fusion.}
Building on the visual captions and Entity Cache from Step 1, we adopt an \textit{asynchronous fusion} strategy to align audio evidence with the existing visual narrative. 
Gemini-2.5-Flash~\cite{comanici2025gemini} processes each corresponding 1-minute audio segment conditioned on both sources, producing a unified audio-visual description and selectively updating entity representations with new audio evidence. 
This yields a chronologically aligned, high-density bimodal event catalog for question generation.

\textbf{Step 3: Agentic Question Generation.}
Although the fused captions capture rich content, they do not explicitly organize event boundaries or long-range dependencies, making direct generation prone to shallow or weakly grounded questions. 
We therefore employ a three-stage workflow powered by GPT-5.1~\cite{singh2025openai}: an \textit{Event Segmentation Agent} groups adjacent captions into coherent event blocks, a \textit{Trajectory Proposal Agent} identifies temporally dispersed and entity-linked evidence chains, and a \textit{Question Generation Agent} converts validated trajectories into four-option MCQs,
following three strict and hierarchical generation principles:

\begin{itemize}[leftmargin=*, itemsep=0pt]
    \item \textit{\textbf{Anti-Shortcut Formulation:}} 
    The question stem must not be answerable via general world knowledge or common sense. It must require direct observation of the content named in the trajectory.
    
    \item \textit{\textbf{Trajectory-Grounded Distractors:}} 
    All incorrect options must be semantically plausible and grounded in actual events from the same video, preventing elimination by surface-level dissimilarity.
    
    \item \textit{\textbf{Stylistic Uniformity:}} 
    All options must be comparable in length, syntactic structure, and wording style to prevent guessing via surface-level cues.
\end{itemize}

\begin{figure}[t]
    \centering
    \includegraphics[width=0.96\textwidth]{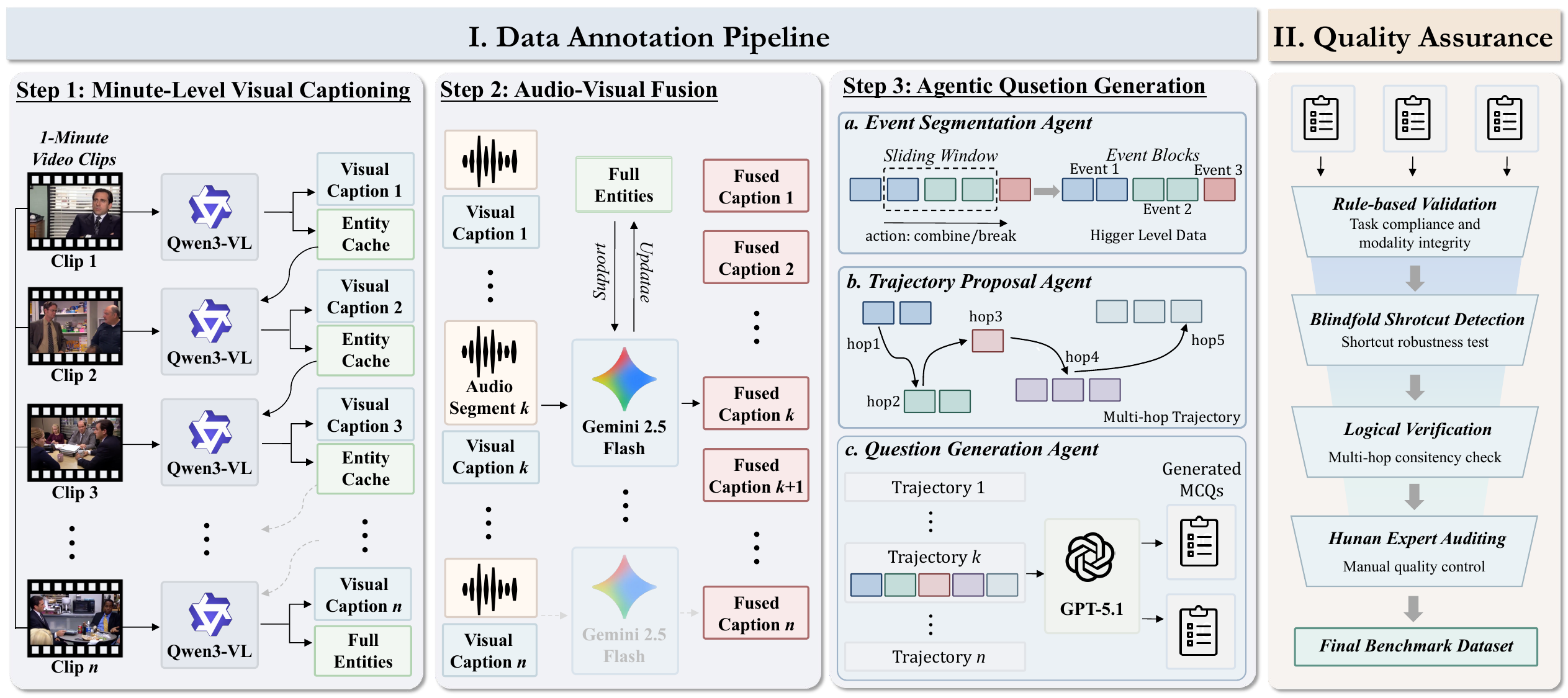}
    \vspace{-1mm}
    \caption{Overview of the \mybench\ data construction pipeline. 
    }
    \label{fig:pipeline}
    \vspace{-4mm}
\end{figure}

\subsection{Quality Assurance}

Despite the strict generation constraints applied in Step 3, the model can still produce questions with latent flaws. 
To ensure the collected questions are high-quality and genuinely multi-hop, we process the generated candidates through a rigorous quality assurance stage led by the following principles:

\begin{itemize}[leftmargin=*, itemsep=0pt]
    \item \textit{\textbf{Rule-based Validation:}} 
    We verify that all questions strictly adhere to their predefined tasks. 
    For example, we enforce that AVR tasks genuinely necessitate evidence from both modalities 
    and that unimodal tasks contain no cross-modal leakage. 
    We additionally verify that each evidence hop is anchored to a specific and verifiable minute-level timestamp and that the MCQ format is strictly adhered to.
    Questions that fail any of these checks are automatically discarded.

    \item \textit{\textbf{Logical Verification:}} 
    A separate LLM conducts a deep audit of each surviving item's multi-hop integrity. 
    It explicitly filters out ``pseudo multi-hop'' questions, checks for answer leakage within the question stem, and ensures that all distractors are semantically relevant and plausibly confusable, avoiding obvious stylistic or length disparities.

    \item \textit{\textbf{Blindfold Shortcut Detection:}} 
    Even with anti-shortcut prompting during generation, some questions may remain answerable through linguistic biases or world knowledge without viewing the video. 
    We therefore submit each item to Gemini~2~Flash with only the question stem and options as text input, without all video content, and record whether the model selects the correct answer.
    Any item answered correctly by this ``blindfolded'' solver is removed from the pool.

    \item \textit{\textbf{Human Expert Auditing:}} 
    As a final quality gate, human annotators manually review a 15\% sample from each generation batch, correcting unclear or weak items and verifying multi-hop integration across the referenced evidence. 
    Any batch whose inspected items exceed a 5\% error or ambiguity rate is discarded. 
    Further auditing details are provided in Appendix~\ref{app:human_audit}.
\end{itemize}
\vspace{-3mm}

\subsection{\mybench\ Statistics}

\textbf{Video statistics.}
The benchmark is built upon \textbf{578} long videos with a total duration of \textbf{339.5 hours}.
Video lengths range from 10.1 to 139.9 minutes, with a mean of 35.2 minutes. The majority (57.6\%) fall in the 30--60 minute range, and 43 videos (7.4\%) exceed one hour, underscoring the ultra-long-form nature of the benchmark.
All videos feature resolutions up to 4K, with 73.7\% at HD (720p) or above, and each contains a stereo audio track encompassing speech, sounds, and music.

\textbf{Question statistics.}
\mybench\ contains \textbf{2,200} questions across four dimensions: Audio-Visual Joint Reasoning (AVR, 7 tasks, 835 questions, 38.0\%), Visual-Centric Reasoning (VR, 2 tasks, 391 questions, 17.8\%), Audio-Centric Reasoning (AR, 3 tasks, 349 questions, 15.9\%), and Multimodal Hallucination (MH, 3 tasks, 625 questions, 28.4\%). 
The benchmark comprises 1,848 single-choice (84.0\%) and 352 multi-choice questions (16.0\%).

\textbf{Multi-hop trajectory statistics.}
Each question is grounded in a multi-hop reasoning chain spanning specific minutes of the video.
On average, each question requires \textbf{3.68} evidence hops, with a temporal span averaging \textbf{15.1 minutes} across the trajectory.

\section{Experiments}

\label{sect:experiments}
\begin{table*}[t]
\caption{
Main results on \mybench\ (\%) across 12 general tasks.
Sub-tasks grouped by three dimensions:
\textbf{AVR} (\textit{Audio-Visual Joint Reasoning}),
\textbf{VR} (\textit{Visual-Centric Reasoning}),
\textbf{AR} (\textit{Audio-Centric Reasoning}).
Abbreviations:
IR = \textit{Information Retrieval},
TS = \textit{Temporal Sequencing},
ET = \textit{Entity Tracking},
FCR = \textit{Forward Causal Reasoning},
BCR = \textit{Backward Causal Reasoning},
CMM = \textit{Cross-Modality Matching},
SL = \textit{Spatiotemporal Localization},
SR = \textit{Spatial Reasoning},
VC = \textit{Visual Counting},
SC = \textit{Speech Context},
ES = \textit{Environmental Sound},
BM = \textit{Background Music}.
\textbf{Avg} is the macro-average over all tasks.
Best result per task is \textbf{bolded}, and best open-source result is \underline{underlined}.
}
\vspace{-1mm}
\label{tab:main_results}
\centering
\renewcommand{\arraystretch}{1.1}
\setlength{\tabcolsep}{4pt}
\begin{adjustbox}{max width=\textwidth}
\begin{tabular}{l *{7}{c} | *{2}{c} | *{3}{c} | c}
\toprule
 & \multicolumn{7}{c|}{\textbf{AVR}} & \multicolumn{2}{c|}{\textbf{VR}} & \multicolumn{3}{c|}{\textbf{AR}} & \\
\cmidrule(lr){2-8}\cmidrule(lr){9-10}\cmidrule(lr){11-13}
\textbf{Model}
  & {\textbf{IR}}
  & {\textbf{TS}}
  & {\textbf{ET}}
  & {\textbf{FCR}}
  & {\textbf{BCR}}
  & {\textbf{CMM}}
  & {\textbf{SL}}
  & {\textbf{SR}}
  & {\textbf{VC}}
  & {\textbf{SC}}
  & {\textbf{ES}}
  & {\textbf{BM}}
  & {\textbf{Avg}} \\
\midrule
\rowcolor{morandiblue}\multicolumn{14}{l}{\textbf{\textit{Closed-source OmniLLMs (With Visual and Audio)}}} \\
\midrule
Gemini 3.1 Pro      & \textbf{83.57} & 60.82          & \textbf{71.77} & \textbf{86.30} & \textbf{61.80} & 49.41          & \textbf{51.54} & \textbf{73.94} & \textbf{41.15} & \textbf{96.92} & 63.64          & \textbf{78.63} & \textbf{68.29} \\
Gemini 2.5 Pro      & 83.57          & \textbf{63.92} & 60.48          & 76.71          & 50.56          & \textbf{54.12} & 48.02          & 68.48          & 39.38          & 83.85          & \textbf{65.91} & 67.18          & 63.52          \\
Gemini 3 Flash      & 82.14          & 53.61          & 65.32          & 83.56          & 59.55          & 49.41          & 29.07          & 73.33          & 36.73          & 86.92          & 61.36          & 66.41          & 62.28          \\
Gemini 2.5 Flash    & 75.00          & 58.76          & 62.10          & 75.34          & 58.43          & 40.00          & 29.07          & 66.06          & 39.38          & 81.54          & 60.23          & 62.60          & 59.04          \\
Gemini 2 Flash      & 66.43          & 53.61          & 58.06          & 64.38          & 43.82          & 41.18          & 25.11          & 54.55          & 27.43          & 70.00          & 55.68          & 58.78          & 51.59          \\
\midrule
\rowcolor{morandiblue}\multicolumn{14}{l}{\textbf{\textit{Open-source OmniLLMs (With Visual and Audio)}}} \\
\midrule
Ming-Flash-Omni-2.0 & \underline{56.43} & \underline{53.61} & \underline{47.58} & \underline{57.53} & 40.45 & \underline{44.71} & 31.28 & 65.45 & \underline{39.38} & 63.85 & 56.82 & \underline{63.36} & \underline{51.70} \\
Qwen3-Omni-30B-A3B  & 47.14 & 51.55 & 35.48 & 43.84 & \underline{50.56} & 40.00 & 32.60 & 58.18 & 38.50 & 63.85 & 59.09 & 60.31 & 48.43 \\
OmniVinci-9B        & 49.29 & 44.33 & 38.71 & \underline{57.53} & 34.83 & 35.29 & 33.48 & 55.15 & 34.51 & \underline{65.38} & \textbf{\underline{65.91}} & 54.20 & 47.38 \\
MiniCPM-o 4.5       & 45.71 & 36.08 & 37.90 & 28.77 & 26.97 & 41.18 & 37.44 & 60.61 & 38.50 & 61.54 & 59.09 & 64.12 & 44.83 \\
Qwen2.5-Omni-7B     & 46.43 & 32.99 & 37.10 & 30.14 & 35.96 & 37.65 & \underline{37.44} & 49.70 & 35.40 & 60.00 & 52.27 & 49.62 & 42.06 \\
Gemma~4-E4B         & 39.29 & 38.14 & 37.10 & 36.99 & 29.21 & 36.47 & 16.74 & 55.15 & 34.07 & 55.38 & 54.55 & 58.02 & 40.93 \\
Video-SALMONN 2     & 42.14 & 41.24 & 29.03 & 30.14 & 29.21 & 32.94 & 31.28 & 48.48 & 39.38 & 47.69 & 47.73 & 44.27 & 38.63 \\
HumanOmni-7B        & 37.86 & 31.96 & 29.84 & 31.51 & 31.46 & 25.88 & 35.68 & 55.15 & 34.96 & 52.31 & 51.14 & 44.27 & 38.50 \\
VideoLLaMA2.1-AV-7B & 36.43 & 29.90 & 25.81 & 17.81 & 16.85 & 25.88 & 22.91 & 38.79 & 38.94 & 36.15 & 39.77 & 35.88 & 30.43 \\
Baichuan-Omni-1.5   & 37.14 & 14.43 & 20.97 & 15.07 & 12.36 & 20.00 & 30.40 & 37.58 & 26.99 & 32.31 & 42.05 & 33.59 & 26.91 \\
\midrule
\rowcolor{morandiblue}\multicolumn{14}{l}{\textbf{\textit{Open-source Single-Modality MLLMs}}} \\
\midrule
Qwen3-VL-32B   & 44.29 & 39.18 & 32.26 & 38.36 & 38.20 & 34.12 & 16.30 & \underline{67.27} & 39.38 & 46.15 & 48.86 & 48.09 & 41.04 \\
Qwen3-VL-8B    & 34.29 & 28.87 & 29.84 & 26.03 & 24.72 & 24.71 & 17.62 & 59.39 & 32.30 & 45.38 & 42.05 & 46.56 & 34.31 \\
Qwen2-Audio-7B & 30.71 & 27.84 & 33.06 & 20.55 & 26.97 & 29.41 & 29.96 & 26.67 & 23.45 & 38.46 & 37.50 & 44.27 & 30.74 \\
\midrule
\rowcolor{morandiblue}\multicolumn{14}{l}{\textbf{\textit{Open-source OmniLLMs (Visual Only)}}} \\
\midrule
Ming-Flash-Omni-2.0 & 37.86 & 35.05 & 32.26 & 45.21 & 31.46 & 27.06 & 19.82 & 54.55 & 38.94 & 43.08 & 47.73 & 49.62 & 38.55 \\
Qwen3-Omni-30B-A3B & 37.86 & 34.02 & 37.90 & 32.88 & 34.83 & 24.71 & 30.84 & 53.94 & 38.94 & 38.46 & 40.91 & 43.51 & 37.40 \\
\midrule
\rowcolor{morandiblue}\multicolumn{14}{l}{\textbf{\textit{Open-source OmniLLMs (Audio Only)}}} \\
\midrule
Ming-Flash-Omni-2.0 & 32.86 & 24.74 & 23.39 & 39.73 & 22.47 & 18.82 & 19.38 & 29.09 & 25.66 & 44.62 & 42.05 & 46.56 & 30.78 \\
Qwen3-Omni-30B-A3B & 27.14 & 31.96 & 20.16 & 23.29 & 33.71 & 29.41 & 18.94 & 27.88 & 21.24 & 43.85 & 40.91 & 45.80 & 30.36 \\
\midrule
Random Guess & 23.39 & 21.39 & 16.33 & 20.21 & 12.64 & 18.53 & 23.57 & 25.00 & 23.78 & 20.58 & 18.75 & 19.27 & 20.29 \\
\bottomrule
\end{tabular}
\end{adjustbox}
\vspace{-3mm}
\end{table*}

\subsection{Experimental Settings}
\label{sect:settings}
\textbf{Evaluated Models.}
We benchmark models in four categories:
\textit{\textbf{1) Closed-source OmniLLMs}}: Gemini~2, Gemini~2.5~\cite{comanici2025gemini} and Gemini~3;
\textit{\textbf{2) Open-source OmniLLMs}}: Qwen3-Omni~\cite{xu2025qwen3}, Qwen2.5-Omni~\cite{xu2025qwen25omnitechnicalreport}, OmniVinci~\cite{ye2025omnivinci}, MiniCPM-o~\cite{yu2025minicpmv45cookingefficient}, HumanOmni~\cite{zhao2025humanomni}, Video-SALMONN~2~\cite{tang2025video}, Baichuan-Omni-1.5~\cite{li2024baichuan}, VideoLLaMA2.1~\cite{cheng2024videollama}, Ming-Flash-Omni~\cite{ai2025ming}, and Gemma~4;
\textit{\textbf{3) Single-modality MLLMs}}: video-only Qwen3-VL~\cite{bai2025qwen3} and audio-only Qwen2-Audio~\cite{chu2024qwen2audio};
\textit{\textbf{4) Visual-only ablations}}: Ming-Flash-Omni and Qwen3-Omni with the audio stream removed.
\textit{\textbf{5) Audio-only ablations}}: Ming-Flash-Omni and Qwen3-Omni with the visual stream removed.

\textbf{Evaluation Protocol.}
For open-source models, we follow official inference configurations and sample as many frames as permitted by the context window to maximize performance. 
For closed-source models, we use the recommended sampling rate (1 frame per second). 
These settings are chosen to obtain the best performance from each evaluated model.
A response is correct only if the predicted option(s) exactly match the ground-truth answer(s).

\begin{table}[t]
\caption{
Hallucination robustness on the MH dimension of \mybench\ (\%).
V2A = \textit{Visual-to-Audio Deception}, A2V = \textit{Audio-to-Visual Deception}, TSF = \textit{Temporal Splicing Fallacy}.
\textbf{MH Avg} is the average over the three MH sub-tasks.
\textbf{Gen.\ Avg} is the average over the 12 general sub-tasks (Table~\ref{tab:main_results}).
Models are sorted by MH Avg within each group.
\textbf{Rank (MH$\to$Gen.)} shows each model's within-group rank by MH Avg and by Gen.\ Avg respectively.
Best result per task is \textbf{bolded}, and best open-source result is \underline{underlined}.
}
\vspace{-1mm}
\label{tab:halluc_results}
\centering
\resizebox{0.85\textwidth}{!}{
\begin{tabular}{lccccc c}
\toprule
\textbf{Model} & \textbf{V2A} & \textbf{A2V} & \textbf{TSF} & \textbf{MH Avg} & \textbf{Gen.\ Avg} & \textbf{Rank (MH$\to$Gen.)} \\
\midrule
\rowcolor{morandiblue}\multicolumn{7}{l}{\textbf{\textit{Closed-source OmniLLMs}}} \\
\midrule
Gemini 3.1 Pro      & \textbf{89.57} & 79.91          & \textbf{84.34} & \textbf{84.61} & \textbf{68.29} & $1 \to 1$ \\
Gemini 3 Flash      & 76.52          & 75.55          & 87.35          & 79.81          & 62.28          & $2 \to 3$ \\
Gemini 2 Flash    & 74.78          & \textbf{81.66} & 77.11          & 77.85          & 51.59          & $ 3\to 5$ \\
Gemini 2.5 Pro      & 79.13          & 74.24          & 75.90          & 76.42          & 63.52          & $4 \to 2$ \\
Gemini 2.5 Flash    & 60.87          & 66.81          & 66.87          & 64.85          & 59.04          & $5 \to 4$ \\
\midrule
\rowcolor{morandiblue}\multicolumn{7}{l}{\textbf{\textit{Open-source OmniLLMs}}} \\
\midrule
Qwen3-Omni-30B-A3B  & 65.65 & 69.87 & \underline{66.87} & \underline{67.46} & 48.43         & $1 \to 2$  \\
Ming-Flash-Omni-2.0 & 71.30 & 67.25 & 62.65 & 67.07            & \underline{51.70} & $2 \to 1$  \\
Gemma~4-E4B         & \underline{74.35} & 69.43 & 56.02 & 66.60            & 40.93          & $3 \to 6$  \\
MiniCPM-o 4.5       & 70.87 & \underline{72.05} & 56.63 & 66.52            & 44.83          & $4 \to 4$  \\
Qwen2.5-Omni-7B     & 60.43 & 55.46 & 53.61 & 56.50            & 42.06          & $5 \to 5$  \\
OmniVinci-9B        & 42.17 & 44.10 & 42.17 & 42.81            & 47.38          & $6 \to 3$  \\
Video-SALMONN 2     & 45.65 & 39.30 & 37.95 & 40.97            & 38.63          & $7 \to 7$  \\
HumanOmni-7B        & 33.91 & 37.99 & 28.31 & 33.40            & 38.50          & $8 \to 8$  \\
Baichuan-Omni-1.5   & 28.26 & 41.48 & 22.29 & 30.68            & 26.91          & $9 \to 10$ \\
VideoLLaMA2.1-AV-7B & 35.22 & 29.69 & 19.88 & 28.26            & 30.43          & $10 \to 9$ \\
\midrule
Random Guess & 21.74 & 22.27 & 17.17 & 20.39 & 20.29 & -- \\
\bottomrule
\end{tabular}
}
\vspace{-2mm}
\end{table}

\subsection{Experimental Findings}

\textbf{Finding 1: Current OmniLLMs still struggle with long-form audio-visual multi-hop reasoning.}
Table~\ref{tab:main_results} shows that \mybench\ remains challenging for all evaluated models.
The strongest closed-source model, Gemini~3.1~Pro, achieves an average accuracy of 68.29\%, while the best open-source model, Ming-Flash-Omni-2.0, reaches 51.70\%.
Closed-source Gemini series models demonstrate \textbf{\textit{dominant performance}}, consistently occupying the top of the rankings,
Nevertheless, even the strongest model remains below 70\%, with particularly limited performance on tasks that require cross-modal matching, spatiotemporal localization, and visual counting.
These results show that \textit{\textbf{current OmniLLMs have not yet mastered the integration of sparse audio-visual evidence across long videos}}.

\textbf{Finding 2: Audio-visual reasoning decomposes into distinct capabilities, with logical synthesis as a major bottleneck.}
The substantial variation across AVR sub-tasks shows that audio-visual reasoning is not a single and uniform capability.
Models perform relatively well when the required information can be directly retrieved or when an observed cause can be mapped to its consequence, but struggle more with cross-modal binding, temporal localization, and backward causal inference.
Gemini~3.1~Pro achieves 83.57\% on Information Retrieval and 86.30\% on Forward Causal Reasoning, while obtaining 49.41\% on Cross-Modality Matching, 51.54\% on Spatiotemporal Localization, and 61.80\% on Backward Causal Reasoning.
Ming-Flash-Omni-2.0 exhibits a similar pattern, reaching 56.43\% on Information Retrieval and 57.53\% on Forward Causal Reasoning, compared with 44.71\% on Cross-Modality Matching, 31.28\% on Spatiotemporal Localization, and 40.45\% on Backward Causal Reasoning.
Taken together, these results suggest that \textit{\textbf{current models can often identify relevant evidence, but remain less reliable when that evidence must be aligned across modalities and composed into a coherent long-range inference}}.

\begin{figure}[t]
  \centering
  \includegraphics[width=\textwidth]{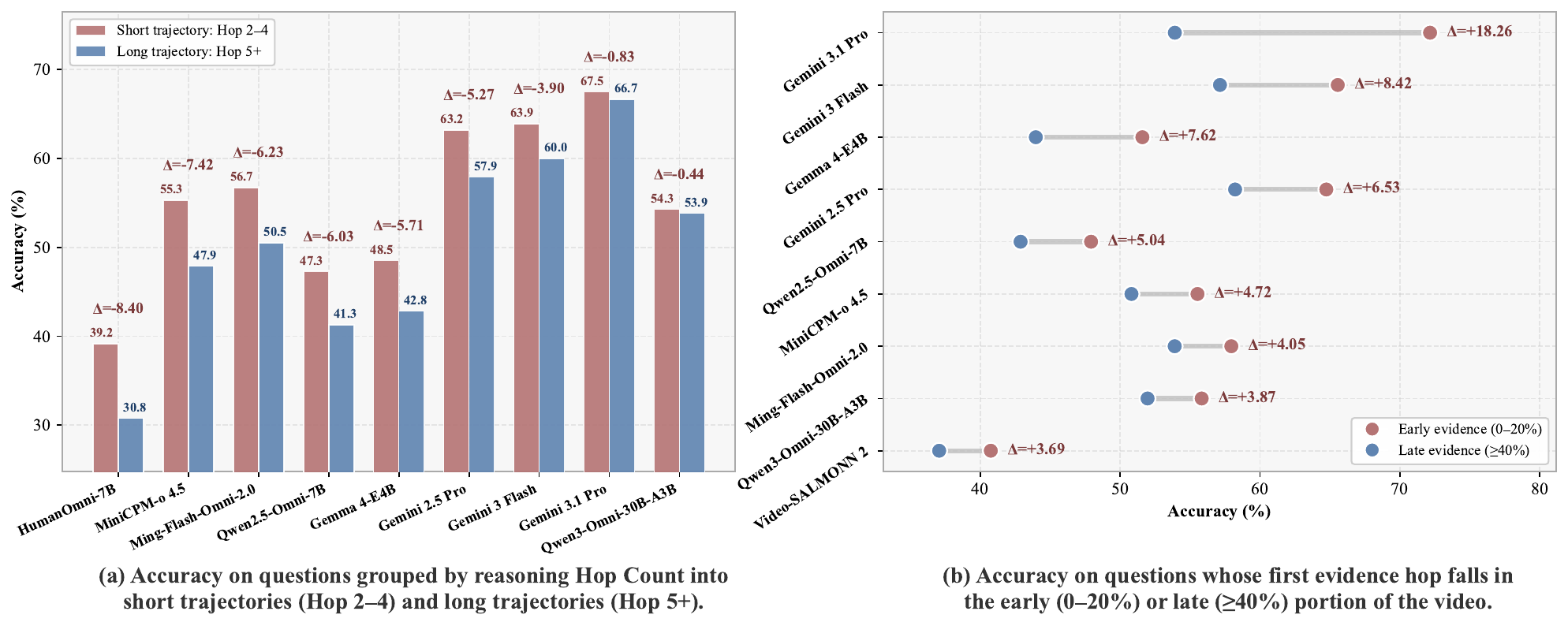}
  \vspace{-1mm}
  \caption{Two diagnostic analyses of OmniLLMs on \mybench.}
  \label{fig:evidence}
  \vspace{-4mm}
\end{figure}

\textbf{Finding 3: Fine-grained visual tracking and non-speech audio understanding remain modality-specific bottlenecks.}
The visual-centric results show that most models perform substantially worse on Visual Counting than on Spatial Reasoning.
Gemini~3.1~Pro achieves 73.94\% on Spatial Reasoning but only 41.15\% on Visual Counting, while Ming-Flash-Omni-2.0 obtains 65.45\% and 39.38\% on the two tasks.
A similar pattern appears for Qwen3-Omni-30B-A3B, whose performance decreases from 58.18\% on Spatial Reasoning to 38.50\% on Visual Counting.
This result indicates that models can often recognize spatial information but struggle to track and accumulate repeated visual events across long videos.
The audio-centric results reveal a related limitation for non-speech information.
For example, Gemini~3.1~Pro achieves 96.92\% on Speech Context, compared with 63.64\% on Environmental Sound and 78.63\% on Background Music.
Across models, performance on Environmental Sound and Background Music is also more variable than performance on Speech Context.
The results indicate that \textit{\textbf{current models reason more reliably about spatial relations and speech content than tracking repeated visual events or interpreting sounds and music throughout long videos}}.

\textbf{Finding 4: General reasoning performance does not reliably reflect hallucination robustness.}
Table~\ref{tab:halluc_results} reveals substantial ranking changes between the general tasks and the Multimodal Hallucination dimension.
Gemma~4 is a representative example, ranking third among open-source models on MH Avg with 66.60\%, while ranking sixth on Gen.\ Avg with 40.93\%.
OmniVinci exhibits the opposite pattern, ranking third on Gen.\ Avg with 47.38\% but sixth on MH Avg with 42.81\%.
A similar shift appears among closed-source models.
Gemini~2~Flash ranks third on MH Avg with 77.85\%, despite ranking fifth on Gen.\ Avg with 51.59\%.
Models also exhibit different strengths across Visual-to-Audio Deception, Audio-to-Visual Deception, and Temporal Splicing Fallacy.
These results show that \textit{\textbf{general audio-visual reasoning performance does not reliably reflect a model's robustness to multimodal hallucinations}}.

\textbf{Finding 5: Longer evidence trajectories are consistently harder to resolve.}
Figure~\ref{fig:evidence}(a) compares questions requiring 2 to 4 evidence hops with those requiring 5 or more hops.
Most evaluated models achieve lower accuracy on the longer trajectories.
MiniCPM-o~4.5 decreases from 55.3\% on shorter trajectories to 47.9\% on longer trajectories, while Ming-Flash-Omni-2.0 decreases from 56.7\% to 50.5\%.
The same pattern appears in the Gemini family, with Gemini~3~Flash decreasing from 63.9\% to 60.0\% and Gemini~3.1~Pro decreasing from 67.5\% to 66.7\%.
Although the magnitude varies across models, the overall trend remains consistent across different architectures.
The consistent degradation with increasing hop count shows that \textit{\textbf{current OmniLLMs struggle with complex long-video reasoning tasks that require aggregating multiple pieces of audio-visual information}}.

\textbf{Finding 6: OmniLLMs favor evidence near the beginning of long videos over evidence in the middle and later portions.}
Figure~\ref{fig:evidence}(b) compares questions whose first evidence hop appears in the first 20\% of the video with questions whose first hop appears at or after 40\%.
Every evaluated model achieves higher accuracy when the first evidence hop appears near the beginning of the video.
On average, accuracy is 6.91\% higher for early-evidence questions, with model-specific differences of 3.69\% to 18.26\%.
Gemini~3.1~Pro exhibits the largest difference of 18.26\%, indicating that sensitivity to evidence position remains notable even in the strongest evaluated model.
These results suggest that \textit{\textbf{current OmniLLMs rely on the opening of a video as a reasoning anchor, which limits their coverage of evidence trajectories that emerge later in the sequence}}.

\section{Conclusion}

In this paper, we present \mybench, the first benchmark to jointly stress \textit{multi-hop trajectory reasoning} and \textit{multimodal hallucination robustness} in the long audio-visual video context.
Built upon a decoupled three-step pipeline and a rigorous multi-stage quality assurance process, \mybench\ comprises 2{,}200 trajectory-grounded multiple-choice questions over 578 long videos. Every question links temporally dispersed evidence, with AVR requiring both audio and vision and VR/AR deliberately isolating visual or auditory reasoning.
We comprehensively evaluate representative OmniLLMs across the general and hallucination dimensions of \mybench, and our analyses surface several insightful findings that we believe offer meaningful guidance for future model training and evaluation.
We hope \mybench\ serves as a rigorous testbed for the next generation of OmniLLMs that reason coherently and faithfully over long, evidence-sparse audio-visual content.

\bibliography{iclr2027_conference}
\bibliographystyle{iclr2027_conference}

\appendix
\section{Error Analysis}

\begin{figure*}[t]
    \centering
    \includegraphics[width=\textwidth]{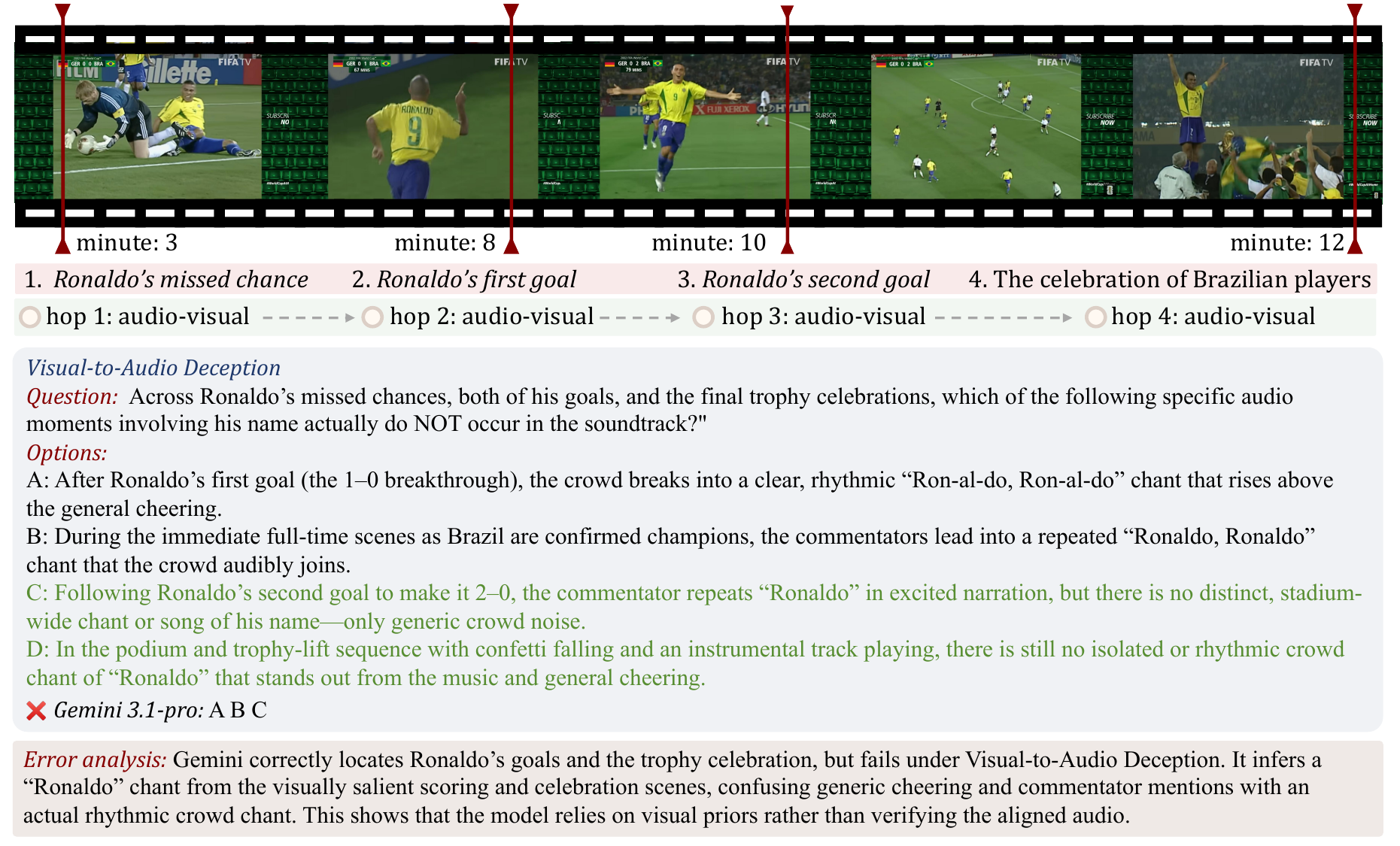}
    \caption{Case of a Visual-to-Audio Deception error. The model infers plausible crowd chants from the visual scenes without verifying whether they occur in the soundtrack.}
    \label{fig:appendix_error_v2a}
\end{figure*}

\begin{figure*}[t]
    \centering
    \includegraphics[width=\textwidth]{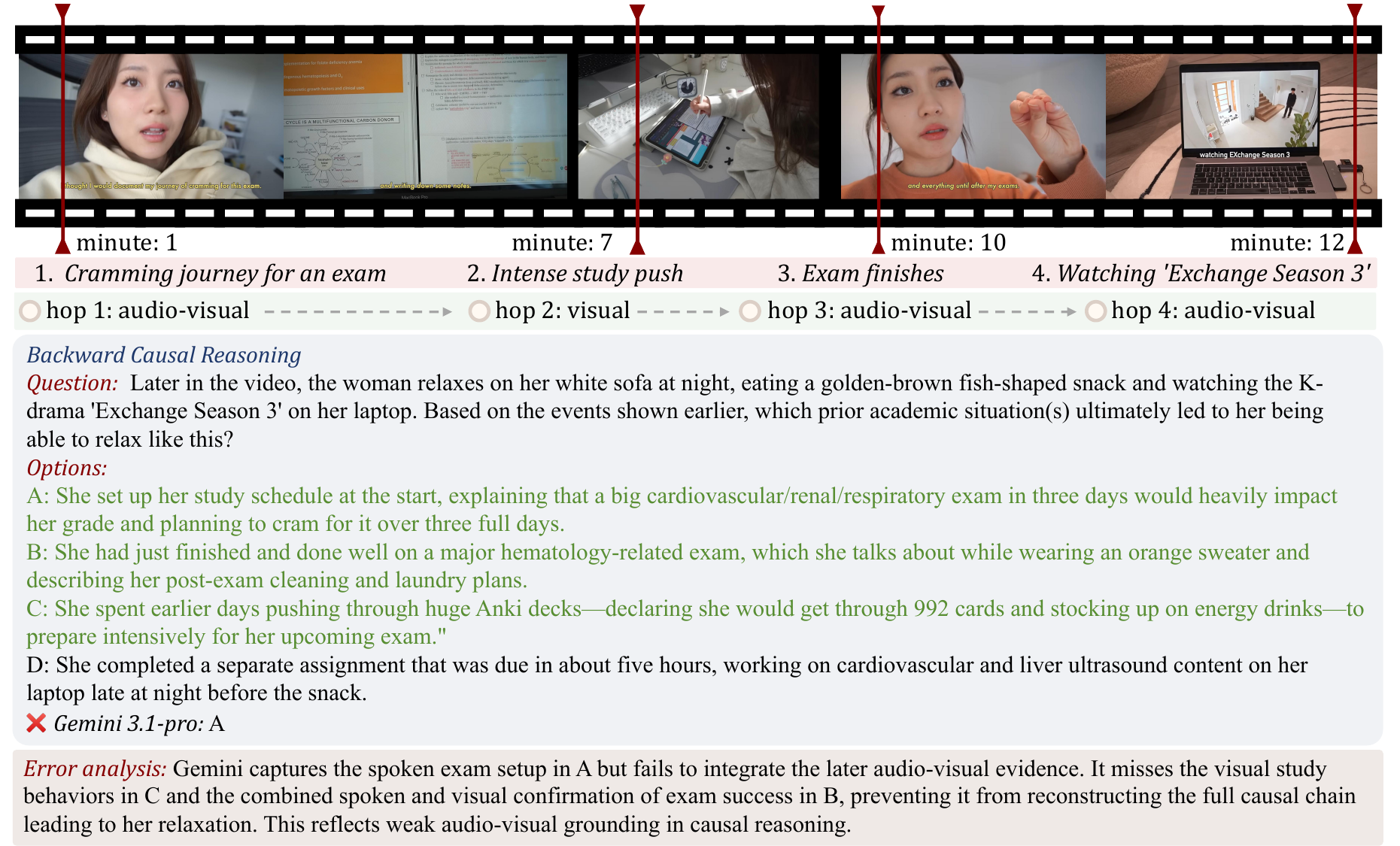}
    \caption{Case of a Backward Causal Reasoning error. The model retrieves partial evidence but fails to reconstruct the complete causal chain across the video.}
    \label{fig:appendix_error_bcr}
\end{figure*}

\begin{figure*}[t]
    \centering
    \includegraphics[width=\textwidth]{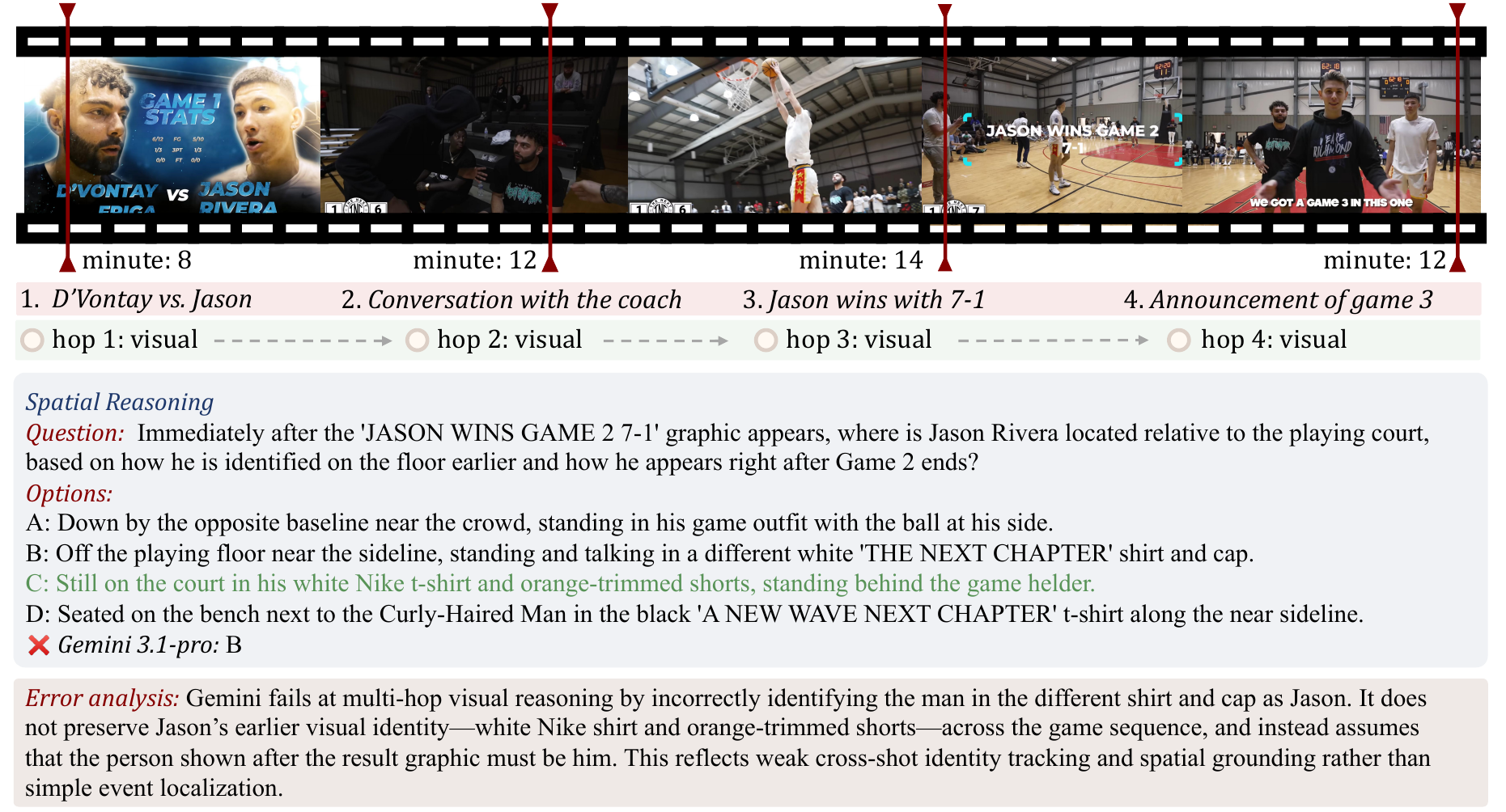}
    \caption{Case of a Spatial Reasoning error. The model fails to preserve entity identity and spatial state across temporally separated shots.}
    \label{fig:appendix_error_spatial}
\end{figure*}

We further examine representative errors to understand why long-context access does not necessarily translate into multi-hop reasoning. We select three cases from Gemini~3.1~Pro, the strongest model in our evaluation, as shown in Figures~\ref{fig:appendix_error_v2a}--\ref{fig:appendix_error_spatial}. 

\paragraph{Common multi-hop failure.}
The three cases share a common structure: the model handles individual moments more successfully than it composes them into a complete trajectory. 
In Figure~\ref{fig:appendix_error_v2a}, the goals and celebrations are visually recognized, but generic cheering and commentator mentions are conflated with a distinct rhythmic crowd chant. 
The model consequently substitutes a cross-modal prior for verified audio evidence. In Figure~\ref{fig:appendix_error_bcr}, one valid antecedent is recovered, yet the later study behavior and examination completion are not jointly incorporated. 
In Figure~\ref{fig:appendix_error_spatial}, the local post-game scene is interpreted without maintaining a consistent identity-location representation across preceding shots. 
Before answering, the model does not reliably ensure that all required hops have been collected, correctly bound to their entities, times, and modalities, and jointly reconciled.

This behavior also illustrates the diagnostic advantage of \mybench. 
Each question is grounded in an explicit, temporally localized, and modality-tagged evidence trajectory. 
A wrong answer can therefore be attributed to missing evidence, incorrect cross-hop binding, or failed aggregation, rather than being reduced to a single undifferentiated accuracy score.
Moreover, the questions cannot be solved from one salient frame, one transcript sentence, or a short temporal neighborhood, and the model must reason over several dispersed events. 

\paragraph{Failure modes across the four dimensions.}

Our broader manual audit reveals distinct and recurring failure patterns beyond the three illustrated examples. 
In \textbf{Audio-Visual Joint Reasoning (AVR)}, a common error is incomplete evidence coverage: models find one valid event and answer before checking the remaining hops, producing only a subset of the correct options. Even when all relevant events are recognized, they may be assigned the wrong temporal order, linked to the wrong entity, or matched across the wrong audio-visual boundary. Similar scenes create strong interference in localization, while causal questions expose poor episode segmentation: models select antecedents that are true and temporally nearby but not causally responsible for the queried outcome. 
In \textbf{Visual-Centric Reasoning (VR)}, models merge distinct but visually similar occurrences, count repeated views as new instances, or miss sparse events near the end of the video. Spatial errors often reflect stale state representations after a scene transition, identity confusion following changes in clothing or viewpoint, and reversal of camera-relative relations such as near versus far. 
In \textbf{Audio-Centric Reasoning (AR)}, models tend to preserve the most salient sound while losing less prominent instances distributed across the trajectory. They also compress a changing soundtrack into an imprecise global summary, confusing the genre, intensity, or narrative role of music across different stages of the video. 
In \textbf{Multimodal Hallucination (MH)}, the dominant behavior is schema-driven completion: models invent visually plausible object attributes, expected crowd chants, or conventional outro speech that never occurs. For temporal splicing, they often verify each fragment independently but fail to test whether the fragments actually form the claimed global chronology or causal narrative. 

The persistence of these errors in the strongest evaluated models shows that the long video audio-visual multihop reasoning capability remains an open challenge.

\section{Evidence-Hop Masking Analysis}
\label{app:hop_masking}

\newcommand{\maskdrop}[2]{\shortstack{#1\\{\scriptsize\textcolor{negative}{(-#2\%)}}}}
\newcommand{\maskgain}[2]{\shortstack{#1\\{\scriptsize\textcolor{positive}{(+#2\%)}}}}

We use evidence-hop masking to validate whether \mybench\ genuinely requires multiple temporally distributed evidence hops.
This validation has three expected outcomes.
First, masking one randomly selected hop (denoted as Mask-1) should reduce accuracy if individual hops contain necessary evidence.
Second, masking all hops (denoted as Mask-All) should cause a further decline if the annotated hops provide complementary information.
Third, performance under Mask-All should approach random guessing if the annotated trajectory captures the evidence required to solve each question.

We evaluate these expectations under three input conditions.
\textit{Full} uses the original audio-visual video.
For \textit{Mask-1}, we randomly select one annotated evidence hop for each question, replace the corresponding video segment with black frames, and mute its audio.
For \textit{Mask-All}, we apply the same operation to every annotated hop, where each hop is mapped to its annotated temporal interval.
When only a minute-level timestamp is available, we use the corresponding one-minute interval, and possible overlapping intervals are merged.

\begin{table*}[t]
\caption{
Evidence-hop masking results on the 12 general sub-tasks of \mybench\ (\%).
\textbf{Full} retains the original audio-visual input, \textbf{Mask-1} removes one randomly selected annotated evidence hop, and \textbf{Mask-All} removes all annotated evidence hops.
Sub-task abbreviations follow Table~\ref{tab:main_results}.
\textbf{Gen. Avg} is the unweighted macro-average over the 12 sub-tasks.
Parenthesized values denote absolute accuracy changes from Full, with red and green indicating degradation and improvement, respectively.
Models follow their order in Table~\ref{tab:main_results}.
}
\label{tab:hop_masking_general}
\centering
\renewcommand{\arraystretch}{1.08}
\setlength{\tabcolsep}{3.5pt}
\begin{adjustbox}{max width=\textwidth}
\begin{tabular}{l *{7}{c} | *{2}{c} | *{3}{c} | c}
\toprule
& \multicolumn{7}{c|}{\textbf{AVR}} & \multicolumn{2}{c|}{\textbf{VR}} & \multicolumn{3}{c|}{\textbf{AR}} & \\
\cmidrule(lr){2-8}\cmidrule(lr){9-10}\cmidrule(lr){11-13}
\textbf{Condition}
  & \textbf{IR} & \textbf{TS} & \textbf{ET} & \textbf{FCR} & \textbf{BCR} & \textbf{CMM} & \textbf{SL}
  & \textbf{SR} & \textbf{VC}
  & \textbf{SC} & \textbf{ES} & \textbf{BM}
  & \textbf{Gen. Avg} \\
\midrule
\rowcolor{morandiblue}\multicolumn{14}{l}{\textbf{\textit{Qwen3-Omni-30B-A3B}}} \\
\midrule
Full     & 47.14 & 51.55 & 35.48 & 43.84 & 50.56 & 40.00 & 32.60 & 58.18 & 38.50 & 63.85 & 59.09 & 60.31 & 48.43 \\
Mask-1   & \maskdrop{26.43}{20.71} & \maskdrop{27.84}{23.71} & \maskdrop{22.58}{12.90} & \maskdrop{23.29}{20.55} & \maskdrop{22.47}{28.09} & \maskdrop{18.82}{21.18} & \maskdrop{23.79}{8.81} & \maskdrop{39.39}{18.79} & \maskdrop{23.89}{14.61} & \maskdrop{44.62}{19.23} & \maskdrop{44.32}{14.77} & \maskdrop{41.98}{18.33} & \maskdrop{29.95}{18.48} \\
Mask-All & \maskdrop{20.00}{27.14} & \maskdrop{23.71}{27.84} & \maskdrop{15.32}{20.16} & \maskdrop{15.07}{28.77} & \maskdrop{12.36}{38.20} & \maskdrop{14.12}{25.88} & \maskdrop{20.70}{11.90} & \maskdrop{29.09}{29.09} & \maskdrop{20.80}{17.70} & \maskdrop{37.69}{26.16} & \maskdrop{34.09}{25.00} & \maskdrop{35.88}{24.43} & \maskdrop{23.24}{25.19} \\
\midrule
\rowcolor{morandiblue}\multicolumn{14}{l}{\textbf{\textit{OmniVinci-9B}}} \\
\midrule
Full     & 49.29 & 44.33 & 38.71 & 57.53 & 34.83 & 35.29 & 33.48 & 55.15 & 34.51 & 65.38 & 65.91 & 54.20 & 47.38 \\
Mask-1   & \maskdrop{44.29}{5.00} & \maskdrop{29.90}{14.43} & \maskdrop{20.16}{18.55} & \maskdrop{35.62}{21.91} & \maskdrop{21.35}{13.48} & \maskdrop{20.00}{15.29} & \maskdrop{23.79}{9.69} & \maskdrop{41.82}{13.33} & \maskdrop{23.89}{10.62} & \maskdrop{46.92}{18.46} & \maskdrop{50.00}{15.91} & \maskdrop{39.69}{14.51} & \maskdrop{33.12}{14.26} \\
Mask-All & \maskdrop{29.29}{20.00} & \maskdrop{17.53}{26.80} & \maskdrop{12.90}{25.81} & \maskdrop{21.92}{35.61} & \maskdrop{12.36}{22.47} & \maskdrop{14.12}{21.17} & \maskdrop{20.70}{12.78} & \maskdrop{32.12}{23.03} & \maskdrop{20.80}{13.71} & \maskdrop{38.46}{26.92} & \maskdrop{40.91}{25.00} & \maskdrop{33.59}{20.61} & \maskdrop{24.56}{22.82} \\
\midrule
\rowcolor{morandiblue}\multicolumn{14}{l}{\textbf{\textit{Gemma~4-E4B}}} \\
\midrule
Full     & 39.29 & 38.14 & 37.10 & 36.99 & 29.21 & 36.47 & 16.74 & 55.15 & 34.07 & 55.38 & 54.55 & 58.02 & 40.93 \\
Mask-1   & \maskdrop{23.57}{15.72} & \maskdrop{22.68}{15.46} & \maskdrop{19.35}{17.75} & \maskdrop{21.92}{15.07} & \maskdrop{16.85}{12.36} & \maskdrop{18.82}{17.65} & \maskgain{23.79}{7.05} & \maskdrop{43.64}{11.51} & \maskdrop{23.89}{10.18} & \maskdrop{39.23}{16.15} & \maskdrop{40.91}{13.64} & \maskdrop{42.75}{15.27} & \maskdrop{28.12}{12.81} \\
Mask-All & \maskdrop{20.00}{19.29} & \maskdrop{17.53}{20.61} & \maskdrop{12.90}{24.20} & \maskdrop{16.44}{20.55} & \maskdrop{10.11}{19.10} & \maskdrop{14.12}{22.35} & \maskgain{20.70}{3.96} & \maskdrop{30.30}{24.85} & \maskdrop{20.80}{13.27} & \maskdrop{30.77}{24.61} & \maskdrop{29.55}{25.00} & \maskdrop{33.59}{24.43} & \maskdrop{21.40}{19.53} \\
\midrule
\rowcolor{morandiblue}\multicolumn{14}{l}{\textbf{\textit{Video-SALMONN~2}}} \\
\midrule
Full     & 42.14 & 41.24 & 29.03 & 30.14 & 29.21 & 32.94 & 31.28 & 48.48 & 39.38 & 47.69 & 47.73 & 44.27 & 38.63 \\
Mask-1   & \maskdrop{25.71}{16.43} & \maskdrop{22.68}{18.56} & \maskdrop{16.13}{12.90} & \maskdrop{20.55}{9.59} & \maskdrop{12.36}{16.85} & \maskdrop{18.82}{14.12} & \maskdrop{23.79}{7.49} & \maskdrop{33.33}{15.15} & \maskdrop{23.89}{15.49} & \maskdrop{35.38}{12.31} & \maskdrop{31.82}{15.91} & \maskdrop{26.72}{17.55} & \maskdrop{24.27}{14.36} \\
Mask-All & \maskdrop{20.00}{22.14} & \maskdrop{17.53}{23.71} & \maskdrop{12.90}{16.13} & \maskdrop{15.07}{15.07} & \maskdrop{8.99}{20.22} & \maskdrop{14.12}{18.82} & \maskdrop{20.70}{10.58} & \maskdrop{24.85}{23.63} & \maskdrop{20.80}{18.58} & \maskdrop{28.46}{19.23} & \maskdrop{23.86}{23.87} & \maskdrop{19.08}{25.19} & \maskdrop{18.86}{19.77} \\
\midrule
Random guess & 23.39 & 21.39 & 16.33 & 20.21 & 12.64 & 18.53
& 23.57 & 25.00 & 23.78 & 20.58 & 18.75 & 19.27
& 20.29 \\
\bottomrule
\end{tabular}
\end{adjustbox}
\end{table*}

\textbf{General tasks validate the multi-hop structure of the benchmark.}
The results satisfy all three expected outcomes.
Averaged across the four evaluated models, Gen. Avg decreases from 43.84\% under Full to 28.87\% under Mask-1 and 22.02\% under Mask-All.
The Mask-All result is close to the 20.29\% accuracy obtained by random guessing.

Masking one hop reduces accuracy in 47 of the 48 combinations of models and sub-tasks.
This consistency shows that a randomly selected annotated hop usually contains information needed to determine the answer.
More importantly, Mask-All underperforms Mask-1 in all 48 combinations.
The remaining hops therefore contribute additional evidence after one hop has already been removed.
This systematic second decline would not be expected if the questions could be solved from a single isolated moment.

Finally, Mask-All reduces the average result from 43.84\% to 22.02\%, which is close to random guessing.
This convergence indicates that the annotated trajectories capture nearly all of the evidence that makes the questions answerable.
Together, the results validate that the general tasks require evidence distributed across multiple annotated hops rather than retrieval from one answer-bearing segment.

\begin{table}[t]
\caption{
Evidence-hop masking results on the Multimodal Hallucination dimension of \mybench\ (\%).
V2A = \textit{Visual-to-Audio Deception}, A2V = \textit{Audio-to-Visual Deception}, and TSF = \textit{Temporal Splicing Fallacy}.
\textbf{MH Avg} is the unweighted macro-average over the three MH sub-tasks.
Parenthesized values denote absolute accuracy degradation from Full.
Models follow their order in Table~\ref{tab:halluc_results}.
}
\label{tab:hop_masking_mh}
\centering
\small
\setlength{\tabcolsep}{5pt}
\renewcommand{\arraystretch}{1.08}
\resizebox{0.44\textwidth}{!}{
\begin{tabular}{lcccc}
\toprule
\textbf{Condition} & \textbf{V2A} & \textbf{A2V} & \textbf{TSF} & \textbf{MH Avg} \\
\midrule
\rowcolor{morandiblue}\multicolumn{5}{l}{\textbf{\textit{Qwen3-Omni-30B-A3B}}} \\
\midrule
Full     & 65.65 & 69.87 & 66.87 & 67.46 \\
Mask-1   & \maskdrop{30.87}{34.78} & \maskdrop{32.31}{37.56} & \maskdrop{27.71}{39.16} & \maskdrop{30.30}{37.16} \\
Mask-All & \maskdrop{23.48}{42.17} & \maskdrop{24.89}{44.98} & \maskdrop{20.48}{46.39} & \maskdrop{22.95}{44.51} \\
\midrule
\rowcolor{morandiblue}\multicolumn{5}{l}{\textbf{\textit{OmniVinci-9B}}} \\
\midrule
Full     & 42.17 & 44.10 & 42.17 & 42.81 \\
Mask-1   & \maskdrop{28.70}{13.47} & \maskdrop{30.57}{13.53} & \maskdrop{30.12}{12.05} & \maskdrop{29.79}{13.02} \\
Mask-All & \maskdrop{19.57}{22.60} & \maskdrop{23.58}{20.52} & \maskdrop{20.48}{21.69} & \maskdrop{21.21}{21.60} \\
\midrule
\rowcolor{morandiblue}\multicolumn{5}{l}{\textbf{\textit{Gemma~4-E4B}}} \\
\midrule
Full     & 74.35 & 69.43 & 56.02 & 66.60 \\
Mask-1   & \maskdrop{28.26}{46.09} & \maskdrop{28.38}{41.05} & \maskdrop{22.29}{33.73} & \maskdrop{26.31}{40.29} \\
Mask-All & \maskdrop{21.74}{52.61} & \maskdrop{21.83}{47.60} & \maskdrop{16.27}{39.75} & \maskdrop{19.95}{46.65} \\
\midrule
\rowcolor{morandiblue}\multicolumn{5}{l}{\textbf{\textit{Video-SALMONN~2}}} \\
\midrule
Full     & 45.65 & 39.30 & 37.95 & 40.97 \\
Mask-1   & \maskdrop{32.17}{13.48} & \maskdrop{24.45}{14.85} & \maskdrop{20.48}{17.47} & \maskdrop{25.70}{15.27} \\
Mask-All & \maskdrop{23.91}{21.74} & \maskdrop{19.65}{19.65} & \maskdrop{15.66}{22.29} & \maskdrop{19.74}{21.23} \\
\midrule
Random guess & 21.74 & 22.27 & 17.17 & 20.39 \\
\bottomrule
\end{tabular}
}
\end{table}

\textbf{Hallucination tasks independently confirm the same multi-hop requirement.}
The MH results exhibit an even clearer progression.
Averaged across the four models, MH Avg decreases from 54.46\% under Full to 28.03\% under Mask-1 and 20.96\% under Mask-All.
The Mask-All result nearly matches the 20.39\% accuracy of random guessing.
Moreover, all 12 combinations of models and sub-tasks follow the same ordering from Full to Mask-1 and then to Mask-All.

These results argue against the possibility that the MH questions can be solved through a generic rejection bias or by verifying only one local cue.
If either shortcut were sufficient, removing the annotated trajectory would not reduce accuracy to the random-guess level.
Instead, the consistent decline from Mask-1 to Mask-All shows that reliable premise verification requires multiple evidence anchors distributed across the audio-visual trajectory.

Taken together, Tables~\ref{tab:hop_masking_general} and~\ref{tab:hop_masking_mh} validate the defining property of \mybench.
Removing one annotated hop consistently impairs prediction.
Removing the remaining hops causes a further decline.
Removing the complete trajectory reduces performance to approximately random guessing in both evaluation settings.
The hops therefore form a cumulative and non-redundant evidence structure that is necessary for solving the benchmark.

\section{Detailed Task Definitions}
\label{sec:appendix_task_def}

This section gives the full definition of each of the fifteen sub-tasks of \mybench, grouped by the four evaluation dimensions introduced in the paper. 
For every sub-task we specify both \textit{what the question asks} and, more importantly, \textit{which capability of OmniLLMs the task is designed to probe}.

\subsection{Audio-Visual Joint Reasoning (AVR)}
The AVR dimension is the conceptual core of \mybench. Every AVR sub-task is constructed so that the answer cannot be recovered from any single modality nor from any single moment in the video. 
Instead, the model must assemble evidence that is simultaneously dispersed across the visual and auditory channels and along the long temporal axis. The seven sub-tasks below probe orthogonal axes of cross-modal multi-hop reasoning.
\begin{itemize}[leftmargin=*, itemsep=2pt]
    \item \textit{\textbf{Information Retrieval (IR):}} 
    IR probes whether a model can retrieve a specific fact (an on-screen number, a spoken name, an embedded text) when the retrieval requires fusing partial cues from at least two distant events, for instance, when the identity of an entity is established at one moment and a fact about that entity is uttered much later. 
    The task therefore stresses long-context cross-modal grounding rather than simply fact recall.
    \item \textit{\textbf{Temporal Sequencing (TS):}} 
    Even when models recover the right set of facts, they often lose the timeline. 
    TS asks the model to chronologically order a small set of events or entity-state changes drawn from non-adjacent positions in the video, isolating the question of whether internal representations preserve temporal order across long spans.
    \item \textit{\textbf{Entity Tracking (ET):}} 
    A persistent challenge for OmniLLMs is re-identifying the same entity after the visual surface changes (clothing, lighting, viewpoint) or the auditory surface changes (different speaker, ambient noise). 
    ET targets the capability that the model must follow a single entity across multiple separate events and reason about its evolving role, relationships, or state.
    \item \textit{\textbf{Forward Causal Reasoning (FCR):}} 
    FCR begins with an early ``cause'' event and asks for its eventual consequence in a much later event. 
    The reasoning chain must traverse intermediate events whose direct relevance is not immediately obvious, isolating the model's ability to project consequences forward across a long timeline.
    \item \textit{\textbf{Backward Causal Reasoning (BCR):}} 
    BCR operates in the reverse direction of FCR. The question is anchored to a late observable outcome and asks for its earlier root cause. We deliberately distinguish between forward and backward causality because the two directions exercise different inference patterns. 
    Backward reasoning, in particular, requires the model to distinguish the true cause from spurious antecedents that merely co-occur along the timeline.
    \item \textit{\textbf{Cross-Modality Matching (CMM):}} 
    CMM probes audio-visual binding under temporal dispersion. 
    The model must pair an auditory cue (a distinctive voice, a sound effect, a musical motif) with its correct visual referent among several visually plausible candidates that appear at different times. 
    The task is designed to expose models that handle each modality competently in isolation but fail to bind them when the evidence is temporally separated.
    \item \textit{\textbf{Spatiotemporal Localization (SL):}} 
    SL asks for the minute-level timestamp at which a specified target event occurs, where the timestamp itself must be triangulated by chaining clues from at least two different events. 
    By forcing the model to commit to an explicit time, the task externalizes whether the model carries a faithful internal clock.
\end{itemize}

\subsection{Visual-Centric Reasoning (VR)}
The VR dimension isolates the visual modality by drawing all evidence from the visual stream alone. 
Its two sub-tasks correspond to the two visual capabilities most commonly degraded by long context:

\begin{itemize}[leftmargin=*, itemsep=2pt]
    \item \textit{\textbf{Spatial Reasoning (SR):}}
    SR asks about spatial relationships (relative positions, directions, layouts) that must hold consistently across multiple events. 
    A correct answer requires the model to maintain a stable mental map of the scene over many minutes of video, rather than re-derive geometry from a single sampled frame.
    \item \textit{\textbf{Visual Counting (VC):}}
    Open-ended counting is a well-known failure mode of MLLMs. 
    VC further conditions the count on a constraint that is itself defined by another event (e.g., \textit{``how many times does action X occur after event Y begins?''}), so that the counting target cannot even be located without first solving an additional reasoning hop. 
\end{itemize}

\subsection{Audio-Centric Reasoning (AR)}
The AR dimension complements VR by isolating the auditory channel. 
Because audio in long videos arrives as a continuous stream rather than as discrete chunks, the three sub-tasks each target a distinct acoustic sub-stream: speech, non-speech environmental sound, and background music.

\begin{itemize}[leftmargin=*, itemsep=2pt]
    \item \textit{\textbf{Speech Context (SC):}}
    SC requires reasoning over spoken dialogue or narration spanning multiple events. 
    To preempt visual leakage, the questions are constructed so that the answer is fully derivable from what is said and not from what is seen, providing a controlled probe of long-form speech comprehension that the visual stream cannot shortcut.
    \item \textit{\textbf{Environmental Sound (ES):}} 
    ES targets non-speech, non-musical sounds such as a siren, a closing door, or machinery. 
    These cues are typically brief and easily missed, and threading them across distant events tests whether the model genuinely attends to acoustic events.
    \item \textit{\textbf{Background Music (BM):}} 
    BM asks about background music or ambient soundscape, including how musical shifts align with the unfolding narrative. 
    The task probes a form of long-horizon audio understanding that carries strong narrative signal in real-world long videos.
\end{itemize}

\subsection{Multimodal Hallucination (MH)}
The MH dimension uses the same exact-match multiple-choice evaluation and measures whether a model can reject a fabricated premise based on the source video. 
Long audio-visual content is a fertile ground for hallucination, because evidence is sparse and the model must commit to an answer under uncertainty.
A model that aces AVR but cannot reject false premises is not yet trustworthy. 

\begin{itemize}[leftmargin=*, itemsep=2pt]
    \item \textit{\textbf{Visual-to-Audio Deception (V2A):}} 
    The question is grounded in genuine visual evidence drawn from the video and then asks about an audio detail that does not occur anywhere in the soundtrack. 
    The intended response is to reject the audio premise rather than confabulate one, which exposes models that over-rely on visual grounding to ``fill in'' the missing audio half.
    \item \textit{\textbf{Audio-to-Visual Deception (A2V):}} 
    A2V is the dual of V2A, where the question is anchored in genuine audio evidence and asks about a visual detail that never actually appears. 
    This direction specifically targets models with strong language priors that synthesize visual content consistent with the spoken context, even when no such visual content was ever observed.
    \item \textit{\textbf{Temporal Splicing Fallacy (TSF):}} 
    TSF presents a fabricated narrative that splices real but temporally isolated fragments into a single ostensibly coherent sequence. 
    The model must explicitly identify the chronological inconsistency, which directly tests whether the model reasons about the global temporal layout of evidence rather than verifying each fragment in isolation.
\end{itemize}

\section{More Detailed Related Work}
\label{app:detailed_related_work}

\subsection{Omnimodal Large Language Models}

Real-world perception is inherently multi-sensory, and closing the gap between human and machine understanding requires models that can handle vision, audio, and language. The research community has broadened the perceptual scope of LLMs, giving rise to Vision-Language Models (VLMs)~\cite{wang2025internvl35, bai2025qwen3, team2026qwen35omni, marafioti2025smolvlm, guo2025seed15vl, liu2025nvila, xiaomi2025mimo}, Audio-Language Models (ALMs)~\cite{tseng2025taste, ding2025kimiaudio, ghosh2025audioflamingo, chu2024qwen2audio, arora2025landscape}, and OmniLLMs~\cite{liu2025ola, luo2025next, fu2024vita, guo2025m2, team2026qwen35omni, tian2026emoomni, ding2026omnisift} that process text, images, videos, and audio within a unified framework. Both open-source models~\cite{ye2025omnivinci, ai2025ming, zhao2025humanomni, tang2025video, xu2025qwen3, xu2025qwen25omnitechnicalreport} and closed-source systems such as the Gemini series~\cite{comanici2025gemini} have shown strong performance across audio and visual tasks.

Recent work further improves fine-grained audio-visual description, temporal alignment, dialogue grounding, robust audio perception, long-form reasoning, and semantic-spatial binding~\cite{ma2026omni, chen2026avocado, tang2026d, yan2026empowering, ghosh2026audio, chen2026scenebind}. These advances provide stronger training data, representations, and model designs for processing extended audio-visual inputs. Agentic long-video systems complement this progress by using iterative search, active observation, structured memory, entity-aware indexing, and tool orchestration to gather task-relevant evidence~\cite{zhang2026deep, wang2025active, rege2026agentic, yin2026hierarchical, tao2025active, xing2026native, xu2026agentic}.

\subsection{MLLM Benchmarks}

MLLM benchmarks have evolved with model capabilities, progressing from modality-isolated suites for image~\cite{liu2024mmbench, yue2024mmmu, yu2023mm} and audio~\cite{sakshi2024mmau, wang2025audiobench, wang2025mmsu, wang2025they} understanding to specialized axes such as fine-grained perception~\cite{ghosh2026understanding, qiang2025ver, weng2026oddgridbench} and multimodal reasoning~\cite{ying2024mmt, yao2025mmreason, du2025easy, li2026mmr, qian2025prism, fu2026videommev2, rawal2024cinepile}. This broad benchmark ecosystem has played an important role in defining model capabilities and exposing systematic failure modes.

The rise of OmniLLMs has led to audio-visual benchmarks covering general perception, cross-modal understanding, reasoning, and model diagnosis~\cite{zhou2025daily, li2024omnibench, chen2026futureomni, cheng2025video, chen2025uno, gong2024av, zhang2025omnieval}. Recent efforts expand this scope through cognitively organized evaluation, long and complex real-world videos, instruction-following captioning, causal analysis of music videos, and evidence-chain data construction~\cite{wang2026avi, goel2026mmou, wang2026omnicap, ghosh2026karma, cai2026omnivideo}. Other benchmarks make valuable contributions to multimodal trustworthiness by testing audio-visual inconsistency and speech-vision hallucination~\cite{chen2026avid, zhang2026svhalluc}. Together, these works provide broad task coverage and increasingly fine-grained evaluation of audio-visual intelligence.

Long-form video evaluation has also advanced rapidly. Existing benchmarks assess long omni-modal understanding~\cite{tao2026lvomnibench, li2025omnivideobench, han2025longinsightbench}, grounded multi-hop reasoning in egocentric videos, movie and television understanding, open-ended movie question answering, multi-timescale comprehension, narrative event structure, and dispersed audio-visual evidence reasoning~\cite{chen2025grounded, ataallah2025infinibench, shaar2026movierecapsqa, ma2026scalelong, asgarov2026nest, xu2026agentic}. These studies establish strong protocols for testing extended temporal context and show that long-video understanding contains several distinct challenges. Building on this foundation, \mybench\ focuses on explicit, temporally dispersed, and modality-tagged multi-hop trajectories, with AVR requiring both audio and vision and VR/AR deliberately isolating one modality. It further places general reasoning and multimodal hallucination evaluation in the same long-form benchmark.

\subsection{More details of the Agentic Question Generation Pipeline}
\label{app:algorithm}

\begin{algorithm}[t]
\IncMargin{1.5em} 
\LinesNotNumbered
\DontPrintSemicolon
\SetKwInOut{Input}{Input}\SetKwInOut{Output}{Output}
\SetKwProg{Stage}{Stage}{:}{}
\SetKwFunction{SegAgent}{SegAgent}
\SetKwFunction{ProposeAgent}{ProposeAgent}
\SetKwFunction{QGenAgent}{QGenAgent}

\caption{Agentic Question Generation Pipeline (Step~3).}
\label{alg:step3}

\Input{Bimodal event catalog $\mathcal{M}=[m_1,\ldots,m_N]$; task type $\tau$; span threshold $\delta$.}
\Output{Released MCQ set $\mathcal{Q}$.}

\Stage{1 -- Event Segmentation Agent}{
  $\mathcal{E}\leftarrow\varnothing$;\ open event $e\leftarrow\{\text{captions}{:}\,[m_1],\ \text{summary}{:}\,m_1\}$\;
  \For{$t = 2$ \KwTo $N$}{
    $(a,\,s')\leftarrow \SegAgent(e.\text{summary},\ m_t,\ m_{t+1},\ m_{t+2})$\;
    \uIf{$a = \textsc{Continue}$}{
      extend $e$ with $m_t$;\ $e.\text{summary}\leftarrow s'$\;
    }
    \Else{
      $\mathcal{E}\leftarrow\mathcal{E}\cup\{e\}$;\ open new event from $m_t$\;
    }
  }
  $\mathcal{E}\leftarrow\mathcal{E}\cup\{e\}$\;
}

\Stage{2 -- Trajectory Proposal Agent}{
  $\mathcal{C}\leftarrow \ProposeAgent(\mathcal{E},\,\tau)$\;
  $\mathcal{T}\leftarrow\bigl\{\,c\in\mathcal{C}\,:\,\text{span}(c)\geq\delta\ \wedge\ \text{adjacent events of }c\text{ share a named entity}\bigr\}$\;
}

\Stage{3 -- Question Generation Agent}{
  $\mathcal{Q}\leftarrow\varnothing$\;
  \For{each trajectory $\mathbf{t}\in\mathcal{T}$}{
    $q\leftarrow \QGenAgent(\mathbf{t},\,\tau)$\;
    refine each per-hop timestamp against the raw minute captions\;
    randomly permute the four answer slots of $q$\;
    $\mathcal{Q}\leftarrow\mathcal{Q}\cup\{q\}$\;
  }
}
\Return $\mathcal{Q}$\;
\end{algorithm}

Algorithm~\ref{alg:step3} consumes the per-minute bimodal catalog $\mathcal{M}$ produced by Steps~1 and~2 and a single span threshold $\delta$ that controls how temporally distant the evidence in a multi-hop chain must be, and it returns the set of trajectory-grounded MCQs $\mathcal{Q}$ that this video contributes to \mybench. We now walk through the three stages in turn.

\textbf{Stage~1: from minutes to events.} 
Directly exposing downstream agents to the entire stream of minute-level captions would force them to reason over dozens of nearly-redundant entries, since adjacent minutes typically belong to the same scene. 
We therefore collapse $\mathcal{M}$ into discrete event blocks via a sliding-window segmentation agent. At each minute~$t$, the agent jointly inspects the running summary of the currently open event, the current minute~$m_t$, and a two-minute look-ahead $(m_{t+1}, m_{t+2})$, and decides whether $m_t$ continues the event or marks a boundary. 
The look-ahead is what distinguishes our segmentation from a purely causal scan. 
When narrative threads briefly interleave (a cutaway, a flashback, a parallel scene), the agent can confirm whether the interruption is incidental or persistent before committing to a boundary, which prevents premature fragmentation of long coherent events. 
The output $\mathcal{E}$ is a partition of the timeline into self-contained event blocks, each carrying its constituent captions, a fused summary, and the union of its entity occurrences.

\textbf{Stage~2: from events to multi-hop trajectories.} 
Operating now on the much shorter sequence $\mathcal{E}$, a trajectory proposal agent generates candidate multi-hop chains together with a per-chain question-direction hint that records the focus, answer cue, distractor cue, and answer-mode preference of the question the chain is intended to support. 
To guarantee that the resulting chains exercise genuine multi-hop reasoning, every candidate is screened by two structural conditions. 
First, its temporal span must exceed $\delta$ event blocks, ruling out trivial chains over consecutive scenes. 
Second, every pair of adjacent events in the chain must share at least one named entity, so that each hop is bridged by explicit coreference rather than by thematic coincidence; this entity-bridge requirement is central to enforcing that no individual event in the chain is alone sufficient to answer the question. 
The chains that survive these conditions form $\mathcal{T}$ and are passed to question generation.

\textbf{Stage~3: from trajectories to MCQs.} 
For each retained trajectory, a question generation agent produces a four-option MCQ following the three generation principles stated in the main paper. 
An anti-shortcut formulation that forces the question to be answerable only by direct observation of the trajectory, trajectory-grounded distractors drawn from other events of the same video, and stylistic uniformity across the four options. 
The agent additionally annotates every hop with a modality label in $\{\text{audio}, \text{video}, \text{audio-video}\}$ and a candidate minute timestamp; the timestamp is then refined by mapping the cited evidence to its closest raw minute caption when it falls outside the event's time range, which yields the fine-grained per-hop grounding used by both the quality-assurance stage and the downstream evaluation. 
Finally, the four answer slots are randomly permuted on a per-question basis so that the position of the correct option is uncorrelated with the generation order, removing any positional bias that models might inherit from its prompting template.

\section{Statistics of \mybench}
\label{app:statistics}

This section provides the detailed statistics of \mybench\. 
Table~\ref{tab:overview_stats} reports the aggregate numbers under this three-axis view, and Table~\ref{tab:benchmark_stats} breaks the question pool down further by sub-task. 

\begin{table}[t]
\caption{Overview statistics of \mybench, organized along three axes: the source video corpus, the released question pool, and the per-question multi-hop reasoning trajectories.}
\label{tab:overview_stats}
\centering
\small
\setlength{\tabcolsep}{6pt}
\resizebox{\textwidth}{!}{
\begin{tabular}{lc|lc|lc}
\toprule
\rowcolor{morandiblue}\multicolumn{2}{l|}{\textbf{Video Corpus}} & \multicolumn{2}{l|}{\textbf{Question Pool}} & \multicolumn{2}{l}{\textbf{Reasoning Trajectories}} \\
\midrule
Total videos        & 578                  & Total questions       & 2{,}200          & Evaluation dimensions    & 4                 \\
Total duration      & 339.5 hrs            & Single-choice         & 1{,}848 (84.0\%) & Sub-tasks                & 15                \\
Duration range      & 10.1--139.9 min      & Multi-choice          &   352 (16.0\%)   & Avg.\ hop count          & 3.68              \\
Avg.\ duration      & 35.2 min             & Avg.\ question length & 43.6 words       & Avg.\ temporal span      & 15.1 min          \\
Resolution          & 73.7\% HD or above   & Avg.\ option length   & 22.1 words       & Modality coverage        & audio + visual    \\
\bottomrule
\end{tabular}
}
\vspace{-2mm}
\end{table}

\textbf{Video corpus.}
The 578 videos in \mybench\ amount to 339.5 hours of continuous audio-visual content, with an average length of 35.2 minutes and a broad range from just over 10 minutes to 2.3 hours. 
The visual quality of the collected videos is high, with 73.7\% of videos at 720p or above and a non-trivial share at 4K. 
The corpus deliberately spans documentaries, vlogs, sports, music performances, and other genres so that no topical shortcut transfers across the benchmark.

\textbf{Question pool.}
The 2{,}200 questions are uniformly four-option multiple-choice items but are not uniformly single-choice: 16.0\% (352 items) are multi-choice, where two to four options can be simultaneously correct. 
Question stems average 43.6 words and options average 22.1 words. 

\textbf{Reasoning trajectories.}
Every question is grounded in an explicit multi-hop trajectory, which is the central structural property that distinguishes \mybench\ from prior audio-visual benchmarks. 
A trajectory comprises 3.68 evidence hops on average and stretches across 15.1 minutes of video, so the typical question requires the model to integrate evidence from roughly four moments separated by a quarter of an hour of footage. 

\textbf{Per-sub-task structure.}
Table~\ref{tab:benchmark_stats} breaks the 2{,}200 items down by the fifteen sub-tasks defined in Appendix~\ref{sec:appendix_task_def}. 
Hop counts and temporal spans vary in interpretable ways across tasks. 
Visual Counting (4.24 hops) and Entity Tracking (4.19 hops) demand the deepest reasoning chains, since both require accumulating evidence over many events, while Information Retrieval (2.87 hops) is structurally the shallowest because the answer hinges on a single key fact assembled from a few cues. 
Spans are widest for Temporal Splicing Fallacy (19.8 min) and Cross-Modality Matching (19.3 min), both of which fundamentally rely on relating widely separated events. 

\begin{table}[t]
\caption{Per-sub-task statistics of \mybench. \textit{Hops}: mean number of evidence hops per question. \textit{Span}: mean temporal gap, in minutes, between the first and last evidence hop. \textit{M-ch}: number of multi-choice items.}
\label{tab:benchmark_stats}
\centering
\footnotesize
\setlength{\tabcolsep}{4pt}
\renewcommand{\arraystretch}{1.05}
\resizebox{0.8\textwidth}{!}{
\begin{tabular}{llcccccc}
\toprule
\rowcolor{morandiblue}\textbf{Dim.} & \textbf{Sub-task (Abbrev.)} & \textbf{\#Q} & \textbf{\%} & \textbf{Hops} & \textbf{Span (min)} & \textbf{M-ch} \\
\midrule
\multirow{7}{*}{\textbf{AVR}}
 & Information Retrieval (IR)         & 140 &  6.4 & 2.87 & 13.5 &  9 \\
 & Temporal Sequencing (TS)           &  97 &  4.4 & 3.99 & 15.2 & 14 \\
 & Entity Tracking (ET)               & 124 &  5.6 & 4.19 & 18.3 & 43 \\
 & Forward Causal Reasoning (FCR)     &  73 &  3.3 & 3.11 & 10.1 & 14 \\
 & Backward Causal Reasoning (BCR)    &  89 &  4.0 & 3.53 & 14.1 & 44 \\
 & Cross-Modality Matching (CMM)      &  85 &  3.9 & 3.80 & 19.3 & 22 \\
 & Spatiotemporal Localization (SL)   & 227 & 10.3 & 3.40 & 12.1 & 13 \\
\midrule
\multirow{2}{*}{\textbf{VR}}
 & Spatial Reasoning (SR)             & 165 &  7.5 & 3.38 & 14.6 &  0 \\
 & Visual Counting (VC)               & 226 & 10.3 & 4.24 & 14.4 & 11 \\
\midrule
\multirow{3}{*}{\textbf{AR}}
 & Speech Context (SC)                & 130 &  5.9 & 3.22 & 15.7 & 23 \\
 & Environmental Sound (ES)           &  88 &  4.0 & 3.41 & 11.8 & 22 \\
 & Background Music (BM)              & 131 &  6.0 & 3.68 & 17.4 & 30 \\
\midrule
\multirow{3}{*}{\textbf{MH}}
 & Visual-to-Audio Deception (V2A)    & 230 & 10.5 & 3.60 & 14.2 & 30 \\
 & Audio-to-Visual Deception (A2V)    & 229 & 10.4 & 4.00 & 15.9 & 25 \\
 & Temporal Splicing Fallacy (TSF)    & 166 &  7.5 & 4.23 & 19.8 & 52 \\
\midrule
\rowcolor{morandiblue}\multicolumn{2}{l}{\textbf{Total}} & \textbf{2{,}200} & \textbf{100.0} & \textbf{3.68} & \textbf{15.1} & \textbf{352} \\
\bottomrule
\end{tabular}
}
\end{table}

\section{Quality Assurance}

\paragraph{Quality assurance yield.}
The 2{,}200 questions retained in \mybench\ are the result of an aggressive multi-stage quality filter applied to a much larger pool of candidate items emitted by the agentic generation pipeline of Section~\ref{app:algorithm}. 
Table~\ref{tab:qa_yield} reports, for each sub-task, the number of items produced before the quality assurance is applied, the number that ultimately enter \mybench, and the corresponding retention rate. 
Across the fifteen sub-tasks, the overall retention rate is 50.5\%. 
Roughly half of the candidate items are dropped during quality assurance. 
The task types that survive at slightly higher rates (localization, counting, and hallucination) are precisely those whose answer spaces are the most structurally constrained, such as integer minutes and integer counts, so a larger fraction of generated candidates already complies with the rule-based portion of the filter and never reaches the more aggressive LLM-based audits.

\begin{figure}[t]
    \centering
    \includegraphics[width=\textwidth]{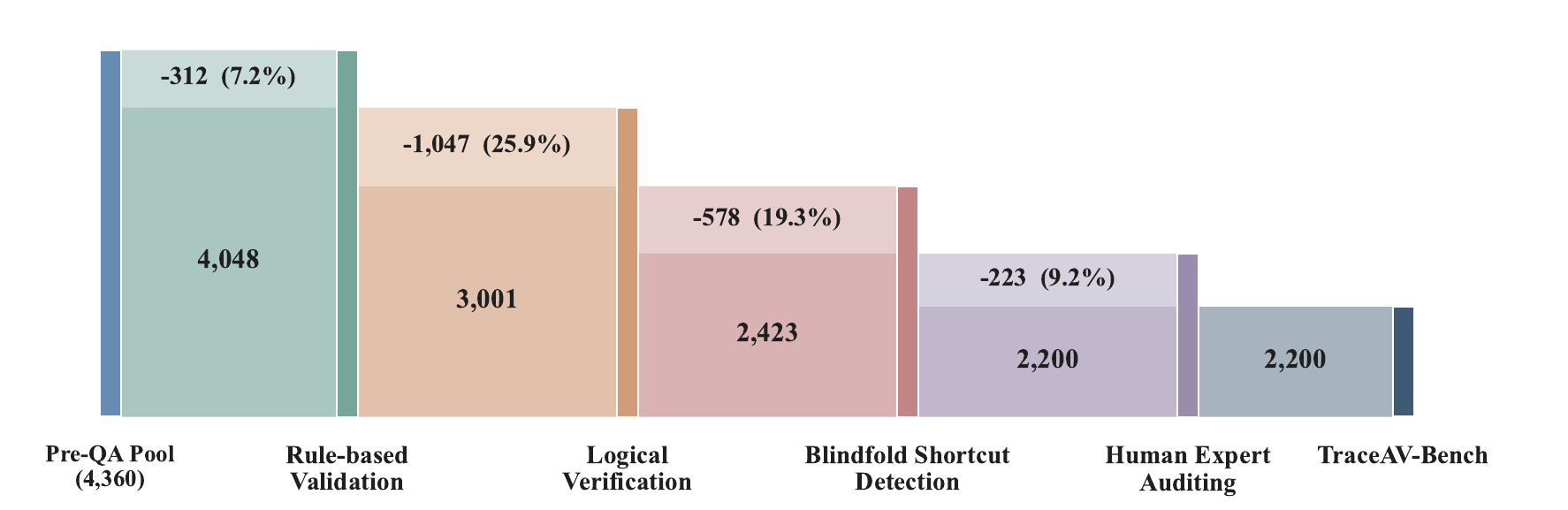}
    \vspace{-2mm}
    \caption{Flow of candidate items through the four-stage quality assurance pipeline.}
    \label{fig:qa_sankey}
    \vspace{-2mm}
\end{figure}

\paragraph{Stage-wise attrition.}
Figure~\ref{fig:qa_sankey} visualizes how candidate items flow through the four sequential QA stages described in Section~\ref{sect:pipeline}. 
Starting from a Pre-QA pool of 4{,}360 items, \textit{Rule-based Validation} removes 312 format- or metadata-violating items, \textit{Logical Verification} prunes the largest share (1{,}047 items) by filtering pseudo multi-hop questions and implausible distractors, \textit{Blindfold Shortcut Detection} eliminates 578 items answerable without the video, and \textit{Human Expert Auditing} discards a final 223 items through reviewer edits and batch rejection, yielding the 2{,}200 questions released in \mybench. 

\begin{table}[t]
\caption{Quality assurance yield per sub-task. \textit{Pre-QA}: number of candidate items produced by the agentic generation pipeline before any quality filter is applied. \textit{Released}: number of items retained in \mybench. \textit{Dropped}: items removed by the quality assurance stage. \textit{Kept}: retention rate, computed as Released / Pre-QA.}
\label{tab:qa_yield}
\centering
\footnotesize
\setlength{\tabcolsep}{5pt}
\renewcommand{\arraystretch}{1.05}
\resizebox{0.85\textwidth}{!}{
\begin{tabular}{llcccc}
\toprule
\rowcolor{morandiblue}\textbf{Dim.} & \textbf{Sub-task (Abbrev.)} & \textbf{Pre-QA} & \textbf{Released} & \textbf{Dropped} & \textbf{Kept (\%)} \\
\midrule
\multirow{7}{*}{\textbf{AVR}}
 & Information Retrieval (IR)         & 280 & 140 & 140 & 50.0 \\
 & Temporal Sequencing (TS)           & 215 &  97 & 118 & 45.1 \\
 & Entity Tracking (ET)               & 265 & 124 & 141 & 46.8 \\
 & Forward Causal Reasoning (FCR)     & 165 &  73 &  92 & 44.2 \\
 & Backward Causal Reasoning (BCR)    & 200 &  89 & 111 & 44.5 \\
 & Cross-Modality Matching (CMM)      & 195 &  85 & 110 & 43.6 \\
 & Spatiotemporal Localization (SL)   & 400 & 227 & 173 & 56.8 \\
\midrule
\multirow{2}{*}{\textbf{VR}}
 & Spatial Reasoning (SR)             & 340 & 165 & 175 & 48.5 \\
 & Visual Counting (VC)               & 400 & 226 & 174 & 56.5 \\
\midrule
\multirow{3}{*}{\textbf{AR}}
 & Speech Context (SC)                & 270 & 130 & 140 & 48.1 \\
 & Environmental Sound (ES)           & 195 &  88 & 107 & 45.1 \\
 & Background Music (BM)              & 280 & 131 & 149 & 46.8 \\
\midrule
\multirow{3}{*}{\textbf{MH}}
 & Visual-to-Audio Deception (V2A)    & 405 & 230 & 175 & 56.8 \\
 & Audio-to-Visual Deception (A2V)    & 400 & 229 & 171 & 57.3 \\
 & Temporal Splicing Fallacy (TSF)    & 350 & 166 & 184 & 47.4 \\
\midrule
\rowcolor{morandiblue}\multicolumn{2}{l}{\textbf{Total}} & \textbf{4{,}360} & \textbf{2{,}200} & \textbf{2{,}160} & \textbf{50.5} \\
\bottomrule
\end{tabular}
}
\end{table}

\section{Human Auditing Protocol and Statistical Reliability}
\label{app:human_audit}

\paragraph{Human role, background, and training.}
Human experts intervened at source-video screening, review of textual intermediates before question generation, final construction-stage quality control, and release auditing.
The release audit was conducted by six graduate-level annotators: four PhD students and two master's students in data science or artificial intelligence.
All annotators had prior research experience in multimodal learning or benchmark construction.
They were native Chinese speakers with professional working proficiency in English. 
Before formal auditing, annotators completed a training session, studied a written decision manual, and independently labeled 30 calibration items.
Calibration disagreements were discussed against the source evidence before the formal audit began.
Each sampled item was independently labeled by three annotators, and a senior PhD annotator who had not supplied an initial label adjudicated disagreements.
Auditors were compensated at USD 15 per hour for training, annotation, and adjudication, consistent with institutional research-assistant rates.

\paragraph{Review of textual intermediates.}
The LLM-generated event catalog was treated as an intermediate representation rather than ground truth.
For every retained video, an author replayed the original audio-video segments and checked the visual captions, fused captions, and entity cache for event coverage, speech and sound fidelity, cross-segment identity consistency, temporal alignment, language errors, and unsupported or sensitive inferences.
Local defects were corrected before question generation, while systematically unreliable catalogs were returned for regeneration.
This pre-generation gate limits error propagation because event segmentation, trajectory proposal, and MCQ generation all depend on these textual intermediates.

\paragraph{Source-grounded release-audit guideline.}
For each sampled item, all three auditors watched every cited segment with its original audio and video, expanding to adjacent context whenever event boundaries or causal relations were unclear.
They formed a source-based judgment before consulting the generated event catalog, MCQ, and trajectory metadata.
An \textit{answer error} meant that the key was unsupported, incomplete, or non-unique; a \textit{timestamp error} meant that the cited interval did not contain the claimed evidence or pointed to the wrong occurrence; and an \textit{unnecessary-hop error} meant that a hop could be removed without making the answer underdetermined.
A \textit{modality-label error} recorded a mismatch between the assigned modality and the evidence actually required; \textit{ambiguous wording} admitted a reasonable answer-changing interpretation; and an \textit{invalid distractor} was duplicated, revealing, contradicted by the stem, or not plausibly confusable from the video.
Reviewers independently assigned \textit{pass}, \textit{modify and retain}, or \textit{reject}.
Fleiss' $\kappa$ was computed from the three pre-adjudication disposition labels, whereas the error and modification rates use adjudicated labels.

\paragraph{Sampling and reliability.}
All 2{,}200 released items have passed the construction pipeline, including its final human quality-control gate at either the inspected-item or batch level; this construction-stage screening is distinct from the release audit below.
The stronger three-reviewer audit covers a fixed sample of 330 candidates from the provisional 2{,}200-item release pool, sampled without replacement within all 15 sub-tasks using proportional allocation with largest-remainder rounding.
The dimension totals were AVR/VR/AR/MH $=126/59/52/93$.
The task allocations, in the order of Table~\ref{tab:benchmark_stats}, were $21/15/19/11/13/13/34$, $25/34$, $19/13/20$, and $34/34/25$, respectively.
For $N=2{,}200$ and $n=330$, the finite-population-corrected worst-case 95\% margin is
\begin{equation}
1.96\sqrt{\frac{0.5(1-0.5)}{330}\frac{2{,}200-330}{2{,}200-1}}
=0.0497,
\end{equation}
or $\pm4.97\%$, and at a 5\% prevalence it is $\pm2.17$ points.
Because the proportional design is approximately self-weighting, hypergeometric inversion gives a finite-population 95\% interval of 2.27--6.41\% for the observed 13/330 item-level error rate (3.9\%).
For reference, zero observed errors would imply a one-sided 95\% upper bound of 18 erroneous items in the population (0.82\%), not proof that every unaudited item is error-free.
The audit therefore quantifies residual-error uncertainty rather than certifying every item.
Table~\ref{tab:audit_results} reports the adjudicated outcomes.
The pre-adjudication Fleiss' $\kappa$ of 0.82 indicates strong agreement among the three independent reviewers.
The four rejected items were removed from the release pool and replaced with reserve items sampled from the same task strata.
Each replacement underwent the same independent three-reviewer source-grounded audit and adjudication procedure before inclusion, so the final benchmark size remains 2{,}200.
The rates in Table~\ref{tab:audit_results} use the original fixed 330-item sample; the replacements were not used to dilute the observed error rate.

\begin{table}[b]
\caption{Adjudicated audit of the fixed 330-item provisional release sample. Error categories are non-exclusive; the item-level row counts each affected item once. The 13 affected items comprise nine corrected and retained items and four rejected items.}
\label{tab:audit_results}
\centering
\footnotesize
\setlength{\tabcolsep}{9pt}
\begin{tabular}{lrr@{\hspace{10pt}}lrr}
\toprule
\rowcolor{morandiblue}\textbf{Outcome/category} & \textbf{Count} & \textbf{Rate} &
\textbf{Outcome/category} & \textbf{Count} & \textbf{Rate} \\
\midrule
Any item-level error & 13 & 3.9\% &
Answer error & 3 & 0.9\% \\
Timestamp error & 5 & 1.5\% &
Unnecessary-hop error & 4 & 1.2\% \\
Modality-label error & 2 & 0.6\% &
Ambiguous wording & 4 & 1.2\% \\
Invalid distractor & 6 & 1.8\% &
Modified and retained & 9 & 2.7\% \\
Rejected after audit & 4 & 1.2\% &
Fleiss' $\kappa$ & \multicolumn{2}{c}{0.82} \\
\bottomrule
\end{tabular}
\end{table}

\paragraph{Construction-stage rejection accounting.}
The preceding automatic stages removed 312/4{,}360 rule-violating items (7.2\%), 1{,}047/4{,}048 logically invalid items (25.9\%), and 578/3{,}001 blindfold-solvable items (19.3\%).
These conditional attrition rates characterize filter behavior, not residual ground-truth accuracy.
A final-gate batch was a generation shard produced from a fixed source-video set and sub-task under one prompt/model version and generation run.
During construction-stage screening, annotators inspected items sampled from every batch, covering 15\% of the 2{,}423 candidates in total; a batch was rejected in full when more than 5\% of its inspected items failed the audit rubric.
Among the 112 batches entering the final human gate (median 21 items, range 12--32), nine exceeded this threshold, removing 171 items through whole-batch rejection.
52 additional candidates were removed through item-level decisions.
Thus, $171+52=223$ candidates were removed at the final stage (9.2\% of 2{,}423), leaving the 2{,}200-item release shown in Figure~\ref{fig:qa_sankey}.
These 223 construction-stage removals are distinct from the four items rejected in the subsequent audit of the provisional release pool.

\section{Additional Experimental Setup}
\label{sect:more_setup}

All open-source models listed are deployed and evaluated on $8\times$NVIDIA H20 GPUs. 
We evaluate the closed-source models using their official APIs. 
We note that most current closed-source multimodal models, such as the GPT series, focus on the visual modality and do not natively accept audio input.
Among the few that genuinely support joint audio-visual input, the Gemini series is the most representative. 
We therefore restrict our closed-source evaluation to the Gemini family. 
The specific versions used are as follows:
\begin{itemize}[leftmargin=*, itemsep=2pt]
    \item Gemini-3.1-Pro-Preview: \texttt{gemini-3.1-pro-preview}
    \item Gemini-3-Flash-Preview: \texttt{gemini-3-flash-preview}
    \item Gemini-2.5-Pro: \texttt{gemini-2.5-pro}
    \item Gemini-2.5-Flash: \texttt{gemini-2.5-flash}
    \item Gemini-2.0-Flash: \texttt{gemini-2.0-flash}
    \item GPT-5.1: \texttt{gpt-5.1-2025-11-13} (for data construction purpose only)
\end{itemize}

\section{More Experimental Findings}
\label{app:more_findings}

\begin{figure}[t]
  \centering
  \includegraphics[width=\textwidth]{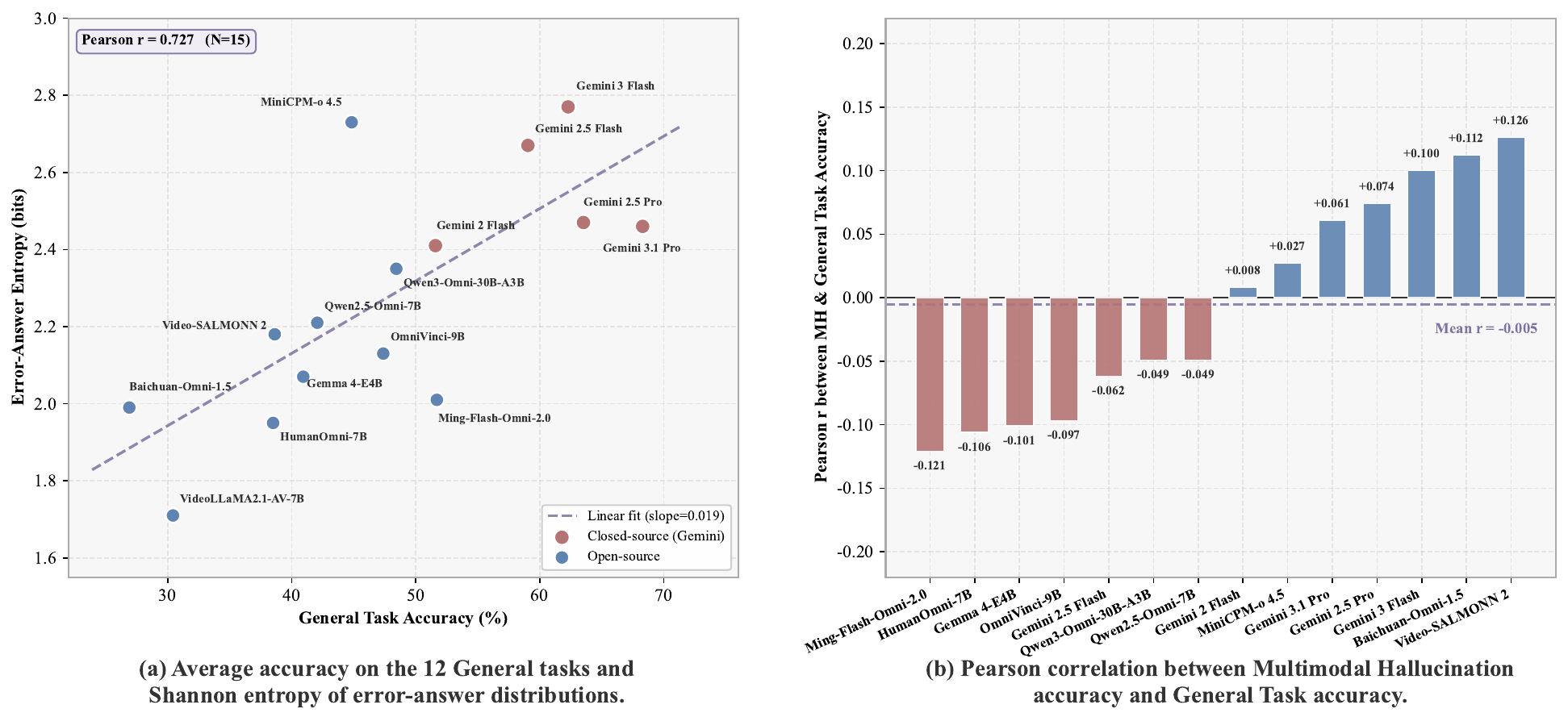}
  \vspace{-3mm}
  \caption{Two diagnostic analyses of OmniLLMs on \mybench.}
  \label{fig:evidence-2}
  \vspace{-5mm}
\end{figure}

\textbf{Stronger models make more \textit{diverse} mistakes.}
Figure~\ref{fig:evidence-2}(a) plots each model's general-task accuracy against the Shannon entropy of its error-answer distribution, $H = -\sum_{i=1}^{k} p_i \log p_i$, where $p_i$ is the frequency of the $i$-th error pattern (the sorted tuple of predicted options) over incorrect predictions.
The two are strongly correlated (Pearson $r{=}$0.727), with weak models such as VideoLLaMA2.1 (30.4\%, 1.71 bits) collapsing errors onto a few patterns and strong models such as Gemini~3~Flash (62.3\%, 2.77 bits) spreading them broadly.
This tight coupling between capability and error diversity suggests that error-answer entropy may reflect a model's potential, offering an indicator during evaluation throughout model training that complements what accuracy alone captures.

\textbf{General-task performance is not a reliable indicator of hallucination robustness.}
Figure~\ref{fig:evidence-2}(b) shows, for each of the 14 models, the Pearson correlation between its Multimodal Hallucination (MH) accuracy and its general-task accuracy computed \textit{across videos}.
All 14 coefficients lie between -0.121 and +0.126, with a mean of $r{=}$-0.005. Thus, a model's performance on the general tasks provides little indication of how robustly it will resist hallucinations on the same video.
This mirrors the rank reversals in Table~\ref{tab:halluc_results}: models that perform better on general tasks are not consistently more robust to multimodal hallucinations.

\section{Evaluation of Agentic Audio-Visual Methods}
\label{app:agentic_evaluation}

Recent agentic methods use active observation, iterative evidence seeking, and tool-mediated memory to reduce the amount of long-video content processed at each reasoning step. We evaluate whether these designs improve multi-hop audio-visual reasoning on \mybench. The evaluation covers three representative methods. \textit{Active Video Perception} (AVP) follows a training-free plan, observe, and reflect workflow for query-conditioned evidence collection~\cite{wang2025active}. \textit{OmniAgent-RL-7B} is a native omni-modal agent trained with agentic supervised fine-tuning and reinforcement learning to alternate between frame, audio, and clip observations~\cite{xing2026native}. \textit{Audio-Guided OmniAgent} uses audio cues to locate candidate events before applying fine-grained visual inspection~\cite{tao2025active}.

\textbf{Evaluation setup.}
AVP uses Qwen3-Omni-30B-A3B as its planner, observer, reflector, and answer synthesizer, with at most three rounds and a confidence threshold of 0.7. Audio-Guided OmniAgent also uses Qwen3-Omni-30B-A3B for its brain and all audio and video tools, with a maximum of 30 iterations. We compare both methods with direct Qwen3-Omni-30B-A3B inference. OmniAgent uses the official OmniAgent-RL-7B checkpoint based on Qwen2.5-Omni-7B, with a maximum of 32 steps. Its matched baseline is direct Qwen2.5-Omni-7B inference. 
Note that the evaluated configurations differ from the default settings of the original methods. 
AVP uses Qwen3-Omni instead of Gemini-2.5-Pro. 
Audio-Guided OmniAgent uses Qwen3-Omni for both the brain and all tools instead of heterogeneous specialized models.

\begin{table*}[t]
\caption{General-task results of agentic audio-visual methods on \mybench\ (\%). General Avg is the macro-average over the 12 sub-tasks. Anchor models repeat representative results from Table~\ref{tab:main_results} for context. AVP and Audio-Guided use direct Qwen3-Omni as the baseline. OmniAgent uses direct Qwen2.5-Omni, its base model.}
\label{tab:agentic_general_results}
\centering
\footnotesize
\setlength{\tabcolsep}{3.5pt}
\renewcommand{\arraystretch}{1.08}
\begin{adjustbox}{max width=\textwidth}
\begin{tabular}{l *{7}{c} | *{2}{c} | *{3}{c} | c}
\toprule
& \multicolumn{7}{c|}{\textbf{AVR}} & \multicolumn{2}{c|}{\textbf{VR}} & \multicolumn{3}{c|}{\textbf{AR}} & \\
\cmidrule(lr){2-8}\cmidrule(lr){9-10}\cmidrule(lr){11-13}
\textbf{Model or method}
& \textbf{IR} & \textbf{TS} & \textbf{ET} & \textbf{FCR} & \textbf{BCR} & \textbf{CMM} & \textbf{SL}
& \textbf{SR} & \textbf{VC}
& \textbf{SC} & \textbf{ES} & \textbf{BM}
& \textbf{Avg} \\
\midrule
Gemini~3.1~Pro & 83.57 & 60.82 & 71.77 & 86.30 & 61.80 & 49.41 & 51.54 & 73.94 & 41.15 & 96.92 & 63.64 & 78.63 & 68.29 \\
Ming-Flash-Omni-2.0 & 56.43 & 53.61 & 47.58 & 57.53 & 40.45 & 44.71 & 31.28 & 65.45 & 39.38 & 63.85 & 56.82 & 63.36 & 51.70 \\
OmniVinci-9B & 49.29 & 44.33 & 38.71 & 57.53 & 34.83 & 35.29 & 33.48 & 55.15 & 34.51 & 65.38 & 65.91 & 54.20 & 47.38 \\
Gemma~4-E4B & 39.29 & 38.14 & 37.10 & 36.99 & 29.21 & 36.47 & 16.74 & 55.15 & 34.07 & 55.38 & 54.55 & 58.02 & 40.93 \\
\midrule
\rowcolor{morandiblue}\multicolumn{14}{l}{\textbf{\textit{Qwen3-Omni-30B-A3B backbone}}} \\
\midrule
Direct & 47.14 & 51.55 & 35.48 & 43.84 & 50.56 & 40.00 & 32.60 & 58.18 & 38.50 & 63.85 & 59.09 & 60.31 & 48.43 \\
AVP & 53.57 & 47.42 & 44.35 & 47.95 & 41.57 & 34.12 & 30.40 & 54.55 & 29.65 & 61.54 & 59.09 & 61.07 & 47.11 \\
Audio-Guided & 50.71 & 41.24 & 38.71 & 53.42 & 41.57 & 27.06 & 18.50 & 44.85 & 27.88 & 67.69 & 45.45 & 51.91 & 42.42 \\
\midrule
\rowcolor{morandiblue}\multicolumn{14}{l}{\textbf{\textit{Qwen2.5-Omni-7B backbone}}} \\
\midrule
Direct & 46.43 & 32.99 & 37.10 & 30.14 & 35.96 & 37.65 & 37.44 & 49.70 & 35.40 & 60.00 & 52.27 & 49.62 & 42.06 \\
OmniAgent-RL-7B & 72.86 & 46.39 & 54.84 & 69.86 & 61.80 & 47.06 & 27.31 & 59.39 & 29.20 & 72.31 & 65.91 & 52.67 & 54.97 \\
\bottomrule
\end{tabular}
\end{adjustbox}
\vspace{-2mm}
\end{table*}

\begin{table}[t]
\caption{Hallucination robustness of agentic audio-visual methods on \mybench\ (\%). MH Avg is the macro-average over V2A, A2V, and TSF. Anchor models repeat representative results from Table~\ref{tab:halluc_results}.}
\label{tab:agentic_halluc_results}
\centering
\small
\setlength{\tabcolsep}{6pt}
\renewcommand{\arraystretch}{1.08}
\resizebox{0.55\textwidth}{!}{
\begin{tabular}{lcccc}
\toprule
\textbf{Model or method} & \textbf{V2A} & \textbf{A2V} & \textbf{TSF} & \textbf{MH Avg} \\
\midrule
Gemini~3.1~Pro & 89.57 & 79.91 & 84.34 & 84.61 \\
Ming-Flash-Omni-2.0 & 71.30 & 67.25 & 62.65 & 67.07 \\
OmniVinci-9B & 42.17 & 44.10 & 42.17 & 42.81 \\
Gemma~4-E4B & 74.35 & 69.43 & 56.02 & 66.60 \\
\midrule
\rowcolor{morandiblue}\multicolumn{5}{l}{\textbf{\textit{Qwen3-Omni-30B-A3B backbone}}} \\
\midrule
Direct & 65.65 & 69.87 & 66.87 & 67.46 \\
AVP & 70.87 & 71.18 & 67.47 & 69.84 \\
Audio-Guided & 60.87 & 50.66 & 62.05 & 57.86 \\
\midrule
\rowcolor{morandiblue}\multicolumn{5}{l}{\textbf{\textit{Qwen2.5-Omni-7B backbone}}} \\
\midrule
Direct & 60.43 & 55.46 & 53.61 & 56.50 \\
OmniAgent-RL-7B & 40.00 & 47.16 & 51.20 & 46.12 \\
\bottomrule
\end{tabular}
}
\vspace{-2mm}
\end{table}

\textbf{Finding 1: Performance gains derived from agentic methods are task-specific and do not transfer uniformly to hallucination robustness.}
OmniAgent achieves the strongest General Avg among the evaluated agentic methods at 54.97\%, improving its direct Qwen2.5-Omni baseline by 12.91\%. The largest gains occur on Information Retrieval, Forward Causal Reasoning, Backward Causal Reasoning, and Entity Tracking. Its performance decreases on Spatiotemporal Localization and Visual Counting. OmniAgent reduces MH Avg by 10.38\% relative to direct Qwen2.5-Omni, with a 20.43\% reduction on Visual-to-Audio Deception.

AVP reaches 47.11\% General Avg, which is 1.32\% below direct Qwen3-Omni. It improves Information Retrieval, Entity Tracking, and Forward Causal Reasoning by 6.43\%, 8.87\%, and 4.11\%. AVP is the only evaluated agentic method that improves all three MH sub-tasks. It reaches 69.84\% MH Avg and gains 2.38\% over its direct baseline. Most of this gain is concentrated in multi-choice questions. AVP increases MH multi-choice accuracy from 18.69\% to 42.06\%, and its MH single-choice accuracy changes from 77.41\% to 75.87\%. Audio-Guided OmniAgent reduces General Avg by 6.01\% and MH Avg by 9.60\% relative to direct Qwen3-Omni.

Agentic gains concentrate on tasks with sparse, identifiable evidence, including retrieval, entity tracking, and forward causal reasoning. 
The same methods remain weak on global counting, precise temporal localization, cross-modal alignment, and premise verification. 

\textbf{Finding 2: Active evidence routing can both fix and introduce errors.}
Compared with direct Qwen3-Omni, AVP corrects 295 previously incorrect answers and introduces 306 new errors. 
Its overall accuracy therefore remains nearly unchanged. 
Audio-Guided OmniAgent corrects 346 errors and introduces 518 new errors, resulting in a clear overall decline. 
These results show that active routing is useful only when it selects the right evidence. 
Searching the wrong video segments leaves the model with incomplete information for the final answer.

\begin{table}[t]
\caption{Inference cost of the three agentic methods over 2{,}200 questions. Mean and P95 wall time are computed from per-sample inference latency.}
\label{tab:agentic_cost}
\centering
\footnotesize
\setlength{\tabcolsep}{4pt}
\renewcommand{\arraystretch}{1.08}
\resizebox{0.54\textwidth}{!}{
\begin{tabular}{lrrr}
\toprule
\textbf{Method} & \textbf{Tokens per item} & \textbf{Mean wall} & \textbf{P95 wall} \\
\midrule
AVP & 39{,}023 & 37.43 s & 86.94 s \\
OmniAgent-RL-7B & 133{,}348 & 308.82 s & 785.69 s \\
Audio-Guided & 200{,}105 & 75.94 s & 244.51 s \\
\bottomrule
\end{tabular}
}
\vspace{-2mm}
\end{table}

\textbf{Finding 3: Longer agent trajectories are associated with higher cost and lower success rates.}
Table~\ref{tab:agentic_cost} shows that AVP has the lowest recorded token count and per-sample latency among the three agentic methods. 
Audio-Guided OmniAgent records 5.13 times as many tokens per item as AVP under the same Qwen3-Omni serving stack. 
Errors also consume more tokens than correct predictions. The increases are 11.8\% for AVP, 49.7\% for OmniAgent, and 17.3\% for Audio-Guided OmniAgent.

The same pattern appears when episodes are grouped by interaction length. 
AVP items that stop after one round achieve 54.49\% accuracy with 33{,}178 tokens on average.
Items reaching its three-round limit achieve 33.88\% with 93{,}102 tokens. 
OmniAgent episodes using 2 to 6 steps achieve 61.10\% with 35{,}243 tokens.
Its 105 episodes reaching the 32-step limit achieve 26.67\% with 458{,}798 tokens. Audio-Guided episodes using at most two high-level actions achieve 50.44\% with 56{,}782 tokens. 
Its 202 episodes reaching the 30-step limit achieve 36.14\% with 823{,}793 tokens. The capped episodes combine high inference cost with low accuracy. In Audio-Guided OmniAgent, 34 capped episodes repeatedly call the same audio QA tool after one global audio caption, showing that repeated tool calls remain a concrete failure mode.

\section{Prompt Templates}
\label{app:prompts}

This section collects every LLM prompt used by the \mybench\ pipeline, grouped by the stage in which it is invoked. 
Placeholder variables filled in at runtime are rendered as \promptvar{variable\_name}. 

\subsection{Pipeline Stage 1: Per-Minute Visual Captioning}
\label{app:prompt_step1}

The visual captioning stage processes each one-minute segment of a video while propagating a running entity cache to keep names and identities consistent across the entire video.

\begin{figure}[htbp]
\begin{tcolorbox}[
  enhanced,
  colback=morandiBack, colframe=morandiTitle,
  title={Stage 1: Visual Captioning with Entity Tracking},
  fonttitle=\bfseries
]
\small\ttfamily
Task: Analyze this 1-minute video segment as a continuous narrative.\\[2pt]
\lbrack Identity Consistency Context\rbrack\\
Refer to these previously identified entities to maintain naming consistency:\\
\promptvar{entity\_cache}\\[2pt]
\lbrack Analysis Requirements\rbrack
\begin{itemize}\setlength{\itemsep}{0pt}
\item Narrative Flow: describe the events as a dynamic sequence, focusing on how actions evolve from start to finish (\promptvar{minute\_summary}).
\item Entity Tracking: (1) identify which entities from the Context above are actively visible or interacting in this specific minute (\promptvar{present\_entities}); (2) if any known entity has changed appearance, state, or role, update its description in \promptvar{present\_entities} to reflect the latest state; (3) identify any NEW prominent people or objects not listed in the Context (\promptvar{new\_entities}).
\item Detail Capture: capture subtle gestures, clothing details, and any on-screen text.
\item Environmental Shift: note changes in setting, lighting, or atmosphere if applicable.
\item For every entity in \promptvar{present\_entities} and \promptvar{new\_entities}, always return a non-empty distinguishing description.
\end{itemize}
\lbrack Strict Constraints\rbrack\\
DO NOT list any entities from the Context that are NOT visible in this segment.\\[2pt]
\lbrack Output Format\rbrack\\
Your response must be a valid JSON object with the keys \promptvar{minute\_summary}, \promptvar{present\_entities}, and \promptvar{new\_entities}.
\end{tcolorbox}
\caption{Prompt used in Stage~1 to generate per-minute visual captions while propagating an entity cache for cross-minute identity consistency.}
\label{prompt:stage1_visual_caption}
\end{figure}

\subsection{Pipeline Stage 2: Audio-Visual Caption Fusion}
\label{app:prompt_step2}

The fusion stage takes the per-minute visual caption together with the corresponding raw audio track and produces a single audio-visual integrated caption while updating entity descriptions with newly observed audio evidence.

\begin{figure}[htbp]
\begin{tcolorbox}[
  enhanced,
  colback=morandiBack, colframe=morandiTitle,
  title={Stage 2: Audio-Visual Caption Fusion},
  fonttitle=\bfseries
]
\small\ttfamily
You are an expert video analyst. Your goal is to synthesize a 1-minute audio-visual integrated caption from two sources: (1)~\lbrack Visual Context\rbrack, a summary of what is seen, and (2)~\lbrack Audio Track\rbrack, the sounds and speech from the same segment.\\[2pt]
\lbrack Visual Context\rbrack: \promptvar{visual\_summary}\\
\lbrack Visual Entities Present\rbrack: \promptvar{visual\_entities}\\[2pt]
\lbrack Analysis Requirements\rbrack
\begin{enumerate}\setlength{\itemsep}{0pt}
\item Information Preservation: the integrated caption MUST encapsulate all information from the visual context.
\item Combined Narrative: integrate the visual description with the actual audio from this segment.
\item Entity Alignment and Evolution: match voices and sounds to the visual entities above when a logical connection exists. If the audio reveals new details about an entity (a spoken name, a distinct voice, a characteristic sound), update its description. If an entity has no relevant audio in this minute, omit it from \promptvar{updated\_entities}.
\item Audio Details: describe background music, environmental sounds, and the dialogue or speech context.
\item Final Output: a single, high-density paragraph that captures both what is seen and what is heard, maintaining chronological and identity consistency.
\end{enumerate}
\lbrack Output Format\rbrack\\
A valid JSON object with keys \promptvar{minute}, \promptvar{integrated\_caption}, and \promptvar{updated\_entities}.
\end{tcolorbox}
\caption{Prompt used in Stage~2 to fuse the per-minute visual caption with the corresponding audio track into a single audio-visual caption.}
\label{prompt:stage2_av_fusion}
\end{figure}

\subsection{Pipeline Stage 3a: Event Segmentation}
\label{app:prompt_step3a}

To convert minute-level captions into discrete events, we slide an LLM judge across the timeline. 
At minute $T$, the model is shown the running summary of the ongoing event, the current minute, and a two-minute look-ahead, and decides whether the current minute continues, transitions out of, or hard-cuts away from the active event.

\begin{figure}[htbp]
\begin{tcolorbox}[
  enhanced,
  colback=morandiBack, colframe=morandiTitle,
  title={Stage 3a: Event Boundary Segmentation},
  fonttitle=\bfseries
]
\small\ttfamily
Task: Analyze the narrative continuity between a past event summary and the current video segment.\\[2pt]
\lbrack Identity Consistency Context\rbrack\\
Active entities in the current minute: \promptvar{current\_entities}\\[2pt]
\lbrack Past State Memory (Ongoing Event)\rbrack: \promptvar{past\_summary}\\
\lbrack Future Vision (Look-ahead Context)\rbrack: T+1 = \promptvar{future\_1}, T+2 = \promptvar{future\_2}\\[2pt]
\lbrack Analysis Requirements\rbrack\\
Determine the relationship of the Current Minute (Minute T) to the Past State Memory:
\begin{itemize}\setlength{\itemsep}{0pt}
\item CONTINUE: direct continuation of narrative and characters.
\item OVERLAP\_TRANSITION: Minute T contains the resolution of the past and the start of the future.
\item HARD\_CUT: total break from the past, aligning instead with the new context in the Future Vision.
\end{itemize}
\lbrack Output Format\rbrack\\
A valid JSON object with keys \promptvar{action}, \promptvar{reason}, and \promptvar{updated\_summary}.
\end{tcolorbox}
\caption{Prompt used in Stage~3a to slide an LLM judge across the timeline and decide CONTINUE / OVERLAP\_TRANSITION / HARD\_CUT at each minute, aggregating contiguous minutes into discrete events.}
\label{prompt:stage3a_segmentation}
\end{figure}

\subsection{Pipeline Stage 3b: Trajectory Proposal}
\label{app:prompt_step3b}

Given a video's event blocks, the trajectory proposer selects multi-hop chains of events that can support a high-quality question for the target task type $\tau$. 
All fifteen task-specific prompts share the generic scaffold shown below.
Note that only the task-specific \textit{Core Principle} differs across tasks, and is listed individually for each task in Figures~\ref{prompt:task_av_retrieval}--\ref{prompt:task_halluc_splicing}.

\begin{figure}[htbp]
\begin{tcolorbox}[
  enhanced,
  colback=morandiBack, colframe=morandiTitle,
  title={Stage 3b: Trajectory Candidate Proposer (Generic Scaffold)},
  fonttitle=\bfseries
]
\small\ttfamily
Task: Select candidate multi-hop trajectories for \promptvar{task\_name} QA.\\[2pt]
\lbrack Core Principle\rbrack\\
\promptvar{task\_specific\_core\_principle} \textit{(per-task body in Figures~\ref{prompt:task_av_retrieval}--\ref{prompt:task_halluc_splicing})}\\[2pt]
\lbrack Trajectory Quality\rbrack\\
Select only HIGH-QUALITY trajectories. A trajectory is high quality when ALL of the following hold:
\begin{enumerate}\setlength{\itemsep}{0pt}
\item STRICT Multi-hop: the intended question MUST be impossible to answer using any single event in the chain, and each hop must be a necessary piece of the puzzle.
\item No Common Sense / No Well-Known Facts: the answer MUST NOT be inferable from general world knowledge or widely known facts, and it must be uniquely tied to specific, non-obvious details of THIS video.
\item Challenging by design: the trajectory must support a question that would genuinely challenge an attentive viewer.
\item Strong evidence: every event in the chain contains concrete, specific evidence (named entity, exact number, quoted speech, visible action).
\item Clear entity bridge: adjacent events share at least one named entity whose state, role, or identity changes meaningfully between them.
\item Reject weak chains: do NOT include trajectories where the connection between events is only thematic or where the answer is guessable without watching the video.
\end{enumerate}
\lbrack Question Direction\rbrack\\
For each candidate, propose a concrete and UNIQUE question direction. Across ALL candidates in this response, no two \promptvar{focus} values may ask about the same fact, entity, or event. The fields \promptvar{focus}, \promptvar{answer\_hint}, \promptvar{distractor\_hint}, and \promptvar{answer\_mode\_hint} (single or multiple, respecting the per-task option-selection mode) must all be supplied.\\[2pt]
\lbrack Output Format\rbrack\\
A valid JSON object containing a list \promptvar{trajectory\_candidates} of items, each with \promptvar{trajectory\_id}, \promptvar{event\_ids}, \promptvar{why\_multihop}, \promptvar{range\_type} (short or long), \promptvar{bridge\_type} (entity, semantic, temporal, or hybrid), and \promptvar{question\_direction}.
\end{tcolorbox}
\caption{Generic scaffold for the Stage~3b trajectory proposer. The same scaffold is instantiated fifteen times, once per task type, by slotting in the task-specific Core Principle listed individually in Figures~\ref{prompt:task_av_retrieval}--\ref{prompt:task_halluc_splicing}.}
\label{prompt:stage3b_proposer}
\end{figure}

The fifteen task-specific instantiations (Core Principle for Stage~3b, Question Design and \promptvar{task\_specific\_key} for Stage~3c) are listed individually in Figures~\ref{prompt:task_av_retrieval}--\ref{prompt:task_halluc_splicing} after the Stage~3c scaffold below.

\subsection{Pipeline Stage 3c: Multiple-Choice Question Generation}
\label{app:prompt_step3c}

Given a selected trajectory, the question generator produces a four-option multiple-choice question, assigns each evidence step a modality label, and writes a per-event minute timestamp. 
The fifteen task-specific prompts again share a common scaffold. 
The variable parts are the \textit{Question Design} block and the \promptvar{task\_specific\_key} pair, both listed per task in Figures~\ref{prompt:task_av_retrieval}--\ref{prompt:task_halluc_splicing}.

\begin{figure}[htbp]
\begin{tcolorbox}[
  enhanced,
  colback=morandiBack, colframe=morandiTitle,
  title={Stage 3c: Multiple-Choice Question Generator (Generic Scaffold)},
  fonttitle=\bfseries
]
\small\ttfamily
Task: Generate one multiple-choice \promptvar{task\_name} question.\\[2pt]
\lbrack Question Design\rbrack\\
\promptvar{task\_specific\_question\_design} \textit{(per-task body in Figures~\ref{prompt:task_av_retrieval}--\ref{prompt:task_halluc_splicing})}\\[2pt]
\lbrack Question Quality Priority\rbrack\\
When generating the question, enforce ALL of the following:
\begin{itemize}\setlength{\itemsep}{0pt}
\item CHALLENGING: genuinely difficult for an attentive viewer; not obvious after a single viewing.
\item NO Common Sense / NO Well-Known Facts: a person who has never seen the video must NOT be able to guess the answer.
\item STRICT Multi-hop: removing even ONE evidence step must make the question unanswerable; every hop must be load-bearing.
\item Strong grounding: every option must be grounded in trajectory evidence; no invented facts.
\item Balanced options: all four options must be plausible on the surface.
\end{itemize}
\lbrack Modality Labeling Rule\rbrack\\
For each evidence step, assign \promptvar{label} based on what is ACTUALLY USED to solve that hop: \texttt{audio} (only auditory cues), \texttt{video} (only visual cues), or \texttt{audio-video} (both modalities jointly required).\\[2pt]
\lbrack Distractor Design Rules\rbrack\\
Distractors MUST satisfy ALL of: (1)~factually grounded in entities, events, numbers, or phrases that actually appear in the video; (2)~logically and commonsensically valid; (3)~superficially plausible to someone who partially watched the video; (4)~length parity with the correct option; (5)~no giveaway language such as ``none of the above''; (6)~no surface-level tells in tone or grammar.\\[2pt]
\lbrack Option Selection Constraint\rbrack\\
Receives a per-task constraint \promptvar{option\_selection\_constraint}: if \texttt{single\_choice\_only}, set \promptvar{question\_type}=``single'' with one item in \promptvar{correct\_options}; if \texttt{single\_or\_multiple}, either single-choice or multiple-choice is allowed (2 to 4 items in \promptvar{correct\_options} for the latter). Always keep exactly four options A/B/C/D.\\[2pt]
\lbrack Timestamp Assignment\rbrack\\
For each node in \promptvar{final\_trajectory}, assign \promptvar{timestamp\_minute} by selecting the single minute within the event's [\promptvar{start\_minute},~\promptvar{end\_minute}] range that best matches the evidence text, using the per-minute \promptvar{raw\_segments} for grounding.\\[2pt]
\lbrack task\_specific\_key\rbrack\\
\promptvar{task\_specific\_key\_pair} \textit{(per-task value in Figures~\ref{prompt:task_av_retrieval}--\ref{prompt:task_halluc_splicing})}\\[2pt]
\lbrack Output Format\rbrack\\
A valid JSON object with keys \promptvar{final\_trajectory} (per-hop \promptvar{event\_id}, \promptvar{evidence}, \promptvar{label}, \promptvar{reason}, \promptvar{timestamp\_minute}), \promptvar{task\_specific\_key}, \promptvar{question\_type}, \promptvar{question}, \promptvar{options} (A/B/C/D), \promptvar{correct\_options}, and \promptvar{answer\_text}.
\end{tcolorbox}
\caption{Generic scaffold for the Stage~3c question generator. The Question Design block and \promptvar{task\_specific\_key} pair are slotted in per task type (Figures~\ref{prompt:task_av_retrieval}--\ref{prompt:task_halluc_splicing}).}
\label{prompt:stage3c_qgen}
\end{figure}

\paragraph{Per-Task Instantiations}

We now list, for each of the fifteen sub-tasks defined in Appendix~\ref{sec:appendix_task_def}, the three pieces that instantiate the generic Stage~3b/3c scaffolds for that task. 
The \textit{Core Principle} that fills the Stage~3b proposer (Figure~\ref{prompt:stage3b_proposer}), the \textit{Question Design} body that fills the Stage~3c generator (Figure~\ref{prompt:stage3c_qgen}), and the \promptvar{task\_specific\_key} pair that the Stage~3c output is required to emit. 
The fifteen figures are ordered to follow the same task taxonomy as Appendix~\ref{sec:appendix_task_def} (AVR $\to$ VR $\to$ AR $\to$ MH).

\begin{figure}[htbp]
\begin{tcolorbox}[
  enhanced,
  colback=morandiBack, colframe=morandiTitle,
  title={Task: Information Retrieval (IR)},
  fonttitle=\bfseries
]
\small\ttfamily
\lbrack Stage 3b Core Principle\rbrack\\
Identify events where a specific key fact (on-screen text, number, spoken name, person identity) is embedded and can only be fully retrieved by chaining evidence from multiple events. Vary the answer position across the trajectory (beginning, middle, or end). Prefer events whose entities carry the attribute ``audio-visual'' so that both modalities contribute to retrieval. Include both short-range and long-range chains.\\[3pt]
\lbrack Stage 3c Question Design\rbrack\\
Identify a specific key fact (on-screen text, number, spoken name, or person identity) as the retrieval target. The question must require chaining at least two evidence steps; the target fact must NOT be directly answerable from a single event. Vary where the answer sits in the chain. Distractors are plausible in-domain values (similar numbers, similar names) drawn from elsewhere in the trajectory.\\[3pt]
\lbrack task\_specific\_key\rbrack\\
\{\promptvar{key\_name}: \texttt{retrieval\_target\_type}, \promptvar{key\_value}: \texttt{text}\,|\,\texttt{number}\,|\,\texttt{speech}\,|\,\texttt{person}\}
\end{tcolorbox}
\caption{Per-task instantiation for \textbf{Information Retrieval (IR)}.}
\label{prompt:task_av_retrieval}
\end{figure}

\begin{figure}[htbp]
\begin{tcolorbox}[
  enhanced,
  colback=morandiBack, colframe=morandiTitle,
  title={Task: Temporal Sequencing (TS)},
  fonttitle=\bfseries
]
\small\ttfamily
\lbrack Stage 3b Core Principle\rbrack\\
Identify events that form a non-trivial temporal order or entity-state evolution that cannot be inferred from any single event. Prefer events where an entity's appearance, role, or state visibly or audibly changes between events. Include both short-range and long-range chains. Avoid trajectories where the ordering is trivially recoverable from event IDs alone.\\[3pt]
\lbrack Stage 3c Question Design\rbrack\\
Ask the model to order events, identify what happened before/after a given event, or describe how an entity's state changed across the trajectory. The correct ordering must require evidence from all events. Distractors should be plausible alternative orderings or state sequences.\\[3pt]
\lbrack task\_specific\_key\rbrack\\
\{\promptvar{key\_name}: \texttt{sequence\_length}, \promptvar{key\_value}: \promptvar{number\_of\_ordered\_items}\}
\end{tcolorbox}
\caption{Per-task instantiation for \textbf{Temporal Sequencing (TS)}.}
\label{prompt:task_av_sequencing}
\end{figure}

\begin{figure}[htbp]
\begin{tcolorbox}[
  enhanced,
  colback=morandiBack, colframe=morandiTitle,
  title={Task: Entity Tracking (ET)},
  fonttitle=\bfseries
]
\small\ttfamily
\lbrack Stage 3b Core Principle\rbrack\\
Centre each trajectory on a specific entity (person, object, or sound source) that appears in at least two non-adjacent events with meaningful state, role, or relationship changes. Leverage entities whose attribute is ``audio-visual'' when both modalities are needed to track the entity. Include both nearby transitions and far-apart isolated transitions. Ensure the question cannot be answered from a single event mention.\\[3pt]
\lbrack Stage 3c Question Design\rbrack\\
Centre the question on a specific entity from the entity pool. Ask about the entity's identity, role, relationship, or state at a specific point in the trajectory, requiring cross-event linking. Distractors should be other entities or plausible but wrong state descriptions drawn from the entity pool or trajectory context.\\[3pt]
\lbrack task\_specific\_key\rbrack\\
\{\promptvar{key\_name}: \texttt{tracked\_entity}, \promptvar{key\_value}: \promptvar{entity\_name}\}
\end{tcolorbox}
\caption{Per-task instantiation for \textbf{Entity Tracking (ET)}.}
\label{prompt:task_av_tracking}
\end{figure}

\begin{figure}[htbp]
\begin{tcolorbox}[
  enhanced,
  colback=morandiBack, colframe=morandiTitle,
  title={Task: Forward Causal Reasoning (FCR)},
  fonttitle=\bfseries
]
\small\ttfamily
\lbrack Stage 3b Core Principle\rbrack\\
Identify event pairs or chains where an earlier event is a clear prerequisite or cause for something that happens in a later event. The causal link must require evidence from both events; neither alone is sufficient to answer ``what happens as a result of X''. Include both local causal chains and distant isolated causal chains.\\[3pt]
\lbrack Stage 3c Question Design\rbrack\\
Frame the question as: ``Given that [cause from early event], what happens / what is the outcome in [later event]?''. The causal link must be non-trivial and require evidence from both events. Distractors should be plausible but causally unrelated or reversed outcomes.\\[3pt]
\lbrack task\_specific\_key\rbrack\\
\{\promptvar{key\_name}: \texttt{causal\_direction}, \promptvar{key\_value}: \texttt{forward}\}
\end{tcolorbox}
\caption{Per-task instantiation for \textbf{Forward Causal Reasoning (FCR)}.}
\label{prompt:task_av_causal_fwd}
\end{figure}

\begin{figure}[htbp]
\begin{tcolorbox}[
  enhanced,
  colback=morandiBack, colframe=morandiTitle,
  title={Task: Backward Causal Reasoning (BCR)},
  fonttitle=\bfseries
]
\small\ttfamily
\lbrack Stage 3b Core Principle\rbrack\\
Identify event pairs or chains where a later event shows an effect or outcome that can only be explained by tracing back to an earlier cause event. The backward causal link must require evidence from both events; neither alone is sufficient to answer ``what caused X''. Include both local causal chains and distant isolated causal chains.\\[3pt]
\lbrack Stage 3c Question Design\rbrack\\
Frame the question as: ``Given that [effect / outcome in later event], what was the cause / prerequisite in [earlier event]?''. The backward causal link must be non-trivial and require evidence from both events. Distractors should be plausible but causally unrelated or forward-direction answers.\\[3pt]
\lbrack task\_specific\_key\rbrack\\
\{\promptvar{key\_name}: \texttt{causal\_direction}, \promptvar{key\_value}: \texttt{backward}\}
\end{tcolorbox}
\caption{Per-task instantiation for \textbf{Backward Causal Reasoning (BCR)}.}
\label{prompt:task_av_causal_bwd}
\end{figure}

\begin{figure}[htbp]
\begin{tcolorbox}[
  enhanced,
  colback=morandiBack, colframe=morandiTitle,
  title={Task: Cross-Modality Matching (CMM)},
  fonttitle=\bfseries
]
\small\ttfamily
\lbrack Stage 3b Core Principle\rbrack\\
Identify events where a specific audio cue (speech phrase, sound, music) and a specific visual element co-occur or are explicitly linked across events. Prefer events containing entities with attribute ``audio-visual''. Also identify events where an audio or visual element appears WITHOUT its expected counterpart, since these support plausible-but-wrong distractors. Include both short-range and long-range chains.\\[3pt]
\lbrack Stage 3c Question Design\rbrack\\
Ask the model to correctly match an audio cue to its visual context (or vice versa) by chaining evidence across events. One option is the correct match; the other three are plausible mismatches (audio from a different event, visually similar but wrong scene). All evidence steps for this task type are labeled \texttt{audio-video}.\\[3pt]
\lbrack task\_specific\_key\rbrack\\
\{\promptvar{key\_name}: \texttt{match\_direction}, \promptvar{key\_value}: \texttt{audio->visual}\,|\,\texttt{visual->audio}\}
\end{tcolorbox}
\caption{Per-task instantiation for \textbf{Cross-Modality Matching (CMM)}.}
\label{prompt:task_av_matching}
\end{figure}

\begin{figure}[htbp]
\begin{tcolorbox}[
  enhanced,
  colback=morandiBack, colframe=morandiTitle,
  title={Task: Spatiotemporal Localization (SL)},
  fonttitle=\bfseries
]
\small\ttfamily
\lbrack Stage 3b Core Principle\rbrack\\
Identify trajectories where pinpointing the EXACT MINUTE of a target event or entity state requires chaining clues from at least two events. Each event is provided with minute-annotated captions \promptvar{minute\_captions} keyed by absolute minute. Prefer multi-minute events so that fine-grained minute-level disambiguation is possible. The question's correct answer and all distractors MUST be concrete minute integers drawn from the trajectory's time range. Include both short-range and long-range chains.\\[3pt]
\lbrack Stage 3c Question Design\rbrack\\
Ask the model to identify the EXACT MINUTE at which a specific event or entity state occurs, requiring evidence from at least two events to narrow it down. The question stem must describe the target event clearly enough that only one minute is correct. ALL FOUR options (A/B/C/D) MUST be concrete integer minute values drawn from the trajectory's time range, NOT descriptive phrases. The correct option is the minute confirmed by the trajectory evidence chain. Distractors should be other plausible minutes from the same trajectory.\\[3pt]
\lbrack task\_specific\_key\rbrack\\
\{\promptvar{key\_name}: \texttt{target\_minute}, \promptvar{key\_value}: \promptvar{correct\_minute\_integer}\}
\end{tcolorbox}
\caption{Per-task instantiation for \textbf{Spatiotemporal Localization (SL)}.}
\label{prompt:task_av_localization}
\end{figure}

\begin{figure}[htbp]
\begin{tcolorbox}[
  enhanced,
  colback=morandiBack, colframe=morandiTitle,
  title={Task: Spatial Reasoning (SR)},
  fonttitle=\bfseries
]
\small\ttfamily
\lbrack Stage 3b Core Principle\rbrack\\
Identify events where spatial relationships (position, direction, relative layout) of objects or persons can only be determined by combining visual evidence from multiple events. Prefer events with rich visual entity descriptions. Audio evidence should NOT be required to answer the question. Include both short-range and long-range chains.\\[3pt]
\lbrack Stage 3c Question Design\rbrack\\
Ask about the spatial relationship (position, direction, relative layout) of objects or persons that can only be determined by combining visual evidence from multiple events. Audio evidence must NOT be needed. Distractors should be plausible but spatially incorrect alternatives. All evidence steps for this task type are labeled \texttt{video}.\\[3pt]
\lbrack task\_specific\_key\rbrack\\
\{\promptvar{key\_name}: \texttt{spatial\_relation\_type},\\
\promptvar{key\_value}: \texttt{position}\,|\,\texttt{direction}\,|\,\texttt{layout}\,|\,\texttt{relative}\}
\end{tcolorbox}
\caption{Per-task instantiation for \textbf{Spatial Reasoning (SR)}.}
\label{prompt:task_v_spatial}
\end{figure}

\begin{figure}[htbp]
\begin{tcolorbox}[
  enhanced,
  colback=morandiBack, colframe=morandiTitle,
  title={Task: Visual Counting (VC)},
  fonttitle=\bfseries
]
\small\ttfamily
\lbrack Stage 3b Core Principle\rbrack\\
Identify trajectories where counting actions, state changes, or objects requires a condition established in one event and the counting target distributed across other events. Example pattern: ``How many times does X do Y after Z happens?''. Visual evidence only; audio should not be required. Include both short-range and long-range chains.\\[3pt]
\lbrack Stage 3c Question Design\rbrack\\
Ask how many times an action occurs, how many objects exist, or how many state changes happen, under a condition established in one event and counted across other events. Produce four distinct integer-count options. Distractors should be nearby integers ($\pm 1, \pm 2$) or counts derived from wrong conditions. Audio evidence must NOT be needed. All evidence steps for this task type are labeled \texttt{video}.\\[3pt]
\lbrack task\_specific\_key\rbrack\\
\{\promptvar{key\_name}: \texttt{count\_target}, \promptvar{key\_value}: \promptvar{what\_is\_being\_counted}\}
\end{tcolorbox}
\caption{Per-task instantiation for \textbf{Visual Counting (VC)}.}
\label{prompt:task_v_counting}
\end{figure}

\begin{figure}[htbp]
\begin{tcolorbox}[
  enhanced,
  colback=morandiBack, colframe=morandiTitle,
  title={Task: Speech Context (SC)},
  fonttitle=\bfseries
]
\small\ttfamily
\lbrack Stage 3b Core Principle\rbrack\\
Identify events where spoken dialogue or narration across multiple events must be combined to answer a question about what was said. Prefer events containing entities with attribute ``audio-visual'' where the speech component is the primary evidence. Visual information should NOT be required. Include both short-range and long-range chains.\\[3pt]
\lbrack Stage 3c Question Design\rbrack\\
Ask about what was said (spoken dialogue or narration) across multiple events, requiring the model to chain speech evidence from at least two events. Visual information must NOT be needed. Distractors should be plausible but incorrect speech content from the same trajectory or nearby events. All evidence steps for this task type are labeled \texttt{audio}.\\[3pt]
\lbrack task\_specific\_key\rbrack\\
\{\promptvar{key\_name}: \texttt{speech\_type}, \promptvar{key\_value}: \texttt{dialogue}\,|\,\texttt{narration}\,|\,\texttt{announcement}\,|\,\texttt{other}\}
\end{tcolorbox}
\caption{Per-task instantiation for \textbf{Speech Context (SC)}.}
\label{prompt:task_a_speech}
\end{figure}

\begin{figure}[htbp]
\begin{tcolorbox}[
  enhanced,
  colback=morandiBack, colframe=morandiTitle,
  title={Task: Environmental Sound (ES)},
  fonttitle=\bfseries
]
\small\ttfamily
\lbrack Stage 3b Core Principle\rbrack\\
Identify events where non-human environmental sounds (rain, machinery, crowd, nature sounds) across multiple events must be combined to answer a question. The question must be answerable from audio alone without visual evidence. Include both short-range and long-range chains.\\[3pt]
\lbrack Stage 3c Question Design\rbrack\\
Ask about non-human environmental sounds (rain, machinery, crowd, nature) that must be identified or reasoned about across multiple events. Visual information must NOT be needed. Distractors should be other plausible environmental sounds from the video. All evidence steps for this task type are labeled \texttt{audio}.\\[3pt]
\lbrack task\_specific\_key\rbrack\\
\{\promptvar{key\_name}: \texttt{sound\_category}, \promptvar{key\_value}: \promptvar{type\_of\_environmental\_sound}\}
\end{tcolorbox}
\caption{Per-task instantiation for \textbf{Environmental Sound (ES)}.}
\label{prompt:task_a_sound}
\end{figure}

\begin{figure}[htbp]
\begin{tcolorbox}[
  enhanced,
  colback=morandiBack, colframe=morandiTitle,
  title={Task: Background Music (BM)},
  fonttitle=\bfseries
]
\small\ttfamily
\lbrack Stage 3b Core Principle\rbrack\\
Identify events where background music, ambient sound, or non-foreground audio cues across multiple events must be combined to answer a question. The question must be answerable from audio alone without visual evidence. Include both short-range and long-range chains.\\[3pt]
\lbrack Stage 3c Question Design\rbrack\\
Ask about background music, ambient sound, or non-foreground audio cues that must be reasoned about across multiple events. Visual information must NOT be needed. Distractors should be other plausible background audio descriptions. All evidence steps for this task type are labeled \texttt{audio}.\\[3pt]
\lbrack task\_specific\_key\rbrack\\
\{\promptvar{key\_name}: \texttt{background\_audio\_type}, \promptvar{key\_value}: \texttt{music}\,|\,\texttt{ambient}\,|\,\texttt{other}\}
\end{tcolorbox}
\caption{Per-task instantiation for \textbf{Background Music (BM)}.}
\label{prompt:task_a_music}
\end{figure}

\begin{figure}[htbp]
\begin{tcolorbox}[
  enhanced,
  colback=morandiBack, colframe=morandiTitle,
  title={Task: Visual-to-Audio Deception (V2A)},
  fonttitle=\bfseries
]
\small\ttfamily
\lbrack Stage 3b Core Principle\rbrack\\
Identify events with rich visual content (entities with attribute ``visual'' or ``audio-visual'') where the visual scene strongly implies an audio detail that is NOT actually present in the video. The trajectory should provide enough visual evidence to make the hallucinated audio seem plausible, while the correct answer is that the audio does not exist. Include both short-range and long-range chains.\\[3pt]
\lbrack Stage 3c Question Design\rbrack\\
Present real visual evidence from the trajectory and ask about an audio detail that does NOT actually exist in the video. The correct answer is the option that states the audio is absent or did not occur. The other three distractors should be fabricated but plausible audio details that the visual scene might suggest. The stem must NOT reveal that this is a hallucination test; phrase it as a genuine audio inquiry. Evidence steps are labeled \texttt{video} (visual cues drive the hallucination).\\[3pt]
\lbrack task\_specific\_key\rbrack\\
\{\promptvar{key\_name}: \texttt{halluc\_type}, \promptvar{key\_value}: \texttt{visual->audio}\}
\end{tcolorbox}
\caption{Per-task instantiation for \textbf{Visual-to-Audio Deception (V2A)}.}
\label{prompt:task_halluc_v2a}
\end{figure}

\begin{figure}[htbp]
\begin{tcolorbox}[
  enhanced,
  colback=morandiBack, colframe=morandiTitle,
  title={Task: Audio-to-Visual Deception (A2V)},
  fonttitle=\bfseries
]
\small\ttfamily
\lbrack Stage 3b Core Principle\rbrack\\
Identify events with rich audio content (entities with attribute ``audio-visual'') where the audio strongly implies a visual detail that is NOT actually present in the video. The trajectory should provide enough audio evidence to make the hallucinated visual seem plausible, while the correct answer is that the visual does not exist. Include both short-range and long-range chains.\\[3pt]
\lbrack Stage 3c Question Design\rbrack\\
Present real audio evidence from the trajectory and ask about a visual detail that does NOT actually exist in the video. The correct answer is the option that states the visual is absent or did not occur. The other three distractors should be fabricated but plausible visual details that the audio might suggest. The stem must NOT reveal that this is a hallucination test; phrase it as a genuine visual inquiry. Evidence steps are labeled \texttt{audio} (audio cues drive the hallucination).\\[3pt]
\lbrack task\_specific\_key\rbrack\\
\{\promptvar{key\_name}: \texttt{halluc\_type}, \promptvar{key\_value}: \texttt{audio->visual}\}
\end{tcolorbox}
\caption{Per-task instantiation for \textbf{Audio-to-Visual Deception (A2V)}.}
\label{prompt:task_halluc_a2v}
\end{figure}

\begin{figure}[htbp]
\begin{tcolorbox}[
  enhanced,
  colback=morandiBack, colframe=morandiTitle,
  title={Task: Temporal Splicing Fallacy (TSF)},
  fonttitle=\bfseries
]
\small\ttfamily
\lbrack Stage 3b Core Principle\rbrack\\
Identify 2 to 4 events from DIFFERENT, non-adjacent time points whose real fragments could be falsely presented as a single coherent sequence. The trajectory should make the splice seem plausible (shared entities or similar settings) while the actual timeline makes the splice impossible. Prefer events with large temporal gaps between them.\\[3pt]
\lbrack Stage 3c Question Design\rbrack\\
Present a fabricated narrative that splices real fragments from different, non-adjacent time points as if they form a coherent sequence. The correct answer must identify that the described sequence is impossible or did not happen as described.
The other three distractors should accept the false narrative or propose alternative but equally false splices. Use the actual timestamps from the trajectory to ground the impossibility.\\[3pt]
\lbrack task\_specific\_key\rbrack\\
\{\promptvar{key\_name}: \texttt{halluc\_type}, \promptvar{key\_value}: \texttt{splice}\}
\end{tcolorbox}
\caption{Per-task instantiation for \textbf{Temporal Splicing Fallacy (TSF)}.}
\label{prompt:task_halluc_splicing}
\end{figure}

\subsection{Quality Assurance Prompts}
\label{app:prompt_qa}

After generation, every candidate item passes through two LLM-based quality checks: a \textit{text-only solver} that flags items whose answer can be guessed without the video, and a \textit{verifier} that audits multi-hop integrity, distractor quality, and answer leakage.

\begin{figure}[htbp]
\begin{tcolorbox}[
  enhanced,
  colback=morandiBack, colframe=morandiTitle,
  title={QA-1: Text-only Direct Solver (System Prompt)},
  fonttitle=\bfseries
]
\small\ttfamily
You are a strict multiple-choice solver. You only see question text and options. You do NOT have access to the video. You are NOT told whether this question is single-choice or multi-choice; infer from wording only. If confidence is insufficient, abstain instead of guessing. Return valid JSON only.
\end{tcolorbox}
\caption{System prompt for the text-only solver, used to detect items whose answer can be guessed from textual cues alone.}
\label{prompt:qa1_solver_sys}
\end{figure}

\begin{figure}[htbp]
\begin{tcolorbox}[
  enhanced,
  colback=morandiBack, colframe=morandiTitle,
  title={QA-1: Text-only Direct Solver (User Prompt)},
  fonttitle=\bfseries
]
\small\ttfamily
Solve the following benchmark question using text-only clues.\\[2pt]
Question: \promptvar{question}\\
Options: A. \promptvar{option\_A}~~B. \promptvar{option\_B}~~C. \promptvar{option\_C}~~D. \promptvar{option\_D}\\[2pt]
Return a JSON object with \promptvar{predicted\_options} (a list of unique letters from A/B/C/D, possibly empty when abstaining), \promptvar{confidence} (a number in $[0,1]$), and \promptvar{reason} (a brief justification).\\[2pt]
Rules: do NOT assume single-choice by default; do NOT force a guess when uncertain; abstain by returning an empty list.
\end{tcolorbox}
\caption{User prompt paired with the text-only solver system prompt in Figure~\ref{prompt:qa1_solver_sys}.}
\label{prompt:qa1_solver_user}
\end{figure}

\begin{figure}[htbp]
\begin{tcolorbox}[
  enhanced,
  colback=morandiBack, colframe=morandiTitle,
  title={QA-2: Item Verifier (System Prompt)},
  fonttitle=\bfseries
]
\small\ttfamily
You are a strict QA benchmark verifier. Given the question, options, reference answer, and trajectory evidence, return JSON only. Evaluate quality and potential flaws conservatively.
\end{tcolorbox}
\caption{System prompt for the item-level verifier that audits multi-hop integrity, distractor quality, and answer leakage.}
\label{prompt:qa2_verifier_sys}
\end{figure}

\begin{figure}[htbp]
\begin{tcolorbox}[
  enhanced,
  colback=morandiBack, colframe=morandiTitle,
  title={QA-2: Item Verifier (User Prompt)},
  fonttitle=\bfseries
]
\small\ttfamily
Verify this benchmark item.\\[2pt]
Task type: \promptvar{task\_type}~~Question type: \promptvar{question\_type}\\
Question: \promptvar{question}\\
Options: A/B/C/D = \promptvar{options}\\
Reference correct\_options: \promptvar{correct\_options}~~Reference answer\_text: \promptvar{answer\_text}\\
Trajectory evidence: \promptvar{trajectory\_with\_timestamps}\\[2pt]
Please evaluate from the following angles:
\begin{enumerate}\setlength{\itemsep}{0pt}
\item Is the multi-hop chain real and correct?
\item Is this pseudo multi-hop (answer solvable by one hop or one local clue)?
\item Option quality: are there obvious superficial signals (e.g., one option much longer or stylistically different); are wrong options relevant rather than random noise; are wrong options plausibly confusing and semantically close?
\item Answer leakage: can the answer be directly inferred from wording in the question stem itself?
\end{enumerate}
Return a JSON object with \promptvar{is\_real\_multihop}, \promptvar{multihop\_issue}, \promptvar{is\_pseudo\_multihop}, \promptvar{pseudo\_multihop\_reason}, \promptvar{option\_quality} (with sub-fields \promptvar{has\_obvious\_pattern}, \promptvar{distractors\_relevant}, \promptvar{distractors\_plausible}, \promptvar{overall\_quality} $\in$ \{\texttt{very\_poor}, \texttt{poor}, \texttt{fair}, \texttt{good}, \texttt{excellent}\}, \promptvar{reason}), \promptvar{answer\_leakage\_detected}, \promptvar{answer\_leakage\_reason}, \promptvar{should\_drop}, \promptvar{drop\_reasons}, \promptvar{severity} $\in$ \{\texttt{none}, \texttt{low}, \texttt{medium}, \texttt{high}\}, and \promptvar{summary}. The verifier sets \promptvar{should\_drop}=\texttt{true} for major quality flaws.
\end{tcolorbox}
\caption{User prompt paired with the verifier system prompt in Figure~\ref{prompt:qa2_verifier_sys}; the structured JSON return is consumed by the dropping rule.}
\label{prompt:qa2_verifier_user}
\end{figure}

\subsection{Evaluation Prompt}
\label{app:prompt_eval}

The same minimal prompt is issued to every candidate model on every \mybench\ item, paired with the original video as input. 
We deliberately keep the wording short and free of chain-of-thought scaffolding 
The model is asked to return only the letter (or comma-separated letters for multi-choice) of its selected option, which makes the response trivially parseable and avoids any prompt-side bias toward verbose explanations.

\begin{figure}[htbp]
\begin{tcolorbox}[
  enhanced,
  colback=morandiBack, colframe=morandiTitle,
  title={Evaluation: Model Inference},
  fonttitle=\bfseries
]
\small\ttfamily
You are given a video. Base your answer on what you see and hear.\\[3pt]
Directly provide the letter representing your choice (A/B/C/D) and nothing else. Do not include the full text of the option; do not provide any explanation. The problem could be a single-choice question or a multiple-choice question. If multiple options are correct, return letters joined by commas (example: A,C).\\[3pt]
Question:\\
\promptvar{question}\\[3pt]
Options:\\
\promptvar{options\_text}
\end{tcolorbox}
\caption{Unified evaluation prompt issued together with the source video to every candidate model on each \mybench\ item.}
\label{prompt:evaluation}
\end{figure}

\newpage

\end{document}